\documentclass[aps,pre,preprint,amsmath,amssymb,superscriptaddress,longbibliography,showkeys]{revtex4-1}

\AtBeginDocument{
 \newwrite\bibnotes
 \def\bibnotesext{Notes.bib}
 \immediate\openout\bibnotes=\jobname\bibnotesext
 \immediate\write\bibnotes{@CONTROL{REVTEX41Control}}
 \immediate\write\bibnotes{@CONTROL{
apsrev41Control,author="08",editor="1",pages="1",title="0",year="1"}}
 \if@filesw
 \immediate\write\@auxout{\string\citation{apsrev41Control}}%
 \fi
}

\usepackage[utf8]{inputenc}
\usepackage{graphicx}
\usepackage{xcolor}
\usepackage[colorlinks=true,citecolor=blue]{hyperref}
\usepackage[caption=false]{subfig}
\usepackage{floatrow}
\usepackage{color}
\usepackage{cancel}
\usepackage{longtable}
\usepackage{multirow}
\usepackage[normalem]{ulem} 

\begin{document}
\raggedbottom
\title{Exploring language relations through syntactic distances and geographic proximity}
\author{Juan De Gregorio}
\author{Ra\'ul Toral}
\author{David S\'anchez}
\affiliation{
Institute for Cross-Disciplinary Physics and Complex Systems IFISC (UIB-CSIC), Campus Universitat de les Illes Balears, E-07122 Palma de Mallorca, Spain
}
\date{\today}
\begin{abstract}
Languages are grouped into families that share common linguistic traits.
While this approach has been successful in understanding genetic relations between diverse languages, more analyses are needed to accurately quantify their relatedness, especially in less studied linguistic levels such as syntax. Here, we explore linguistic distances using series of parts of speech (POS) extracted from the Universal Dependencies dataset. Within an information-theoretic framework, we show that employing POS trigrams maximizes the possibility of capturing syntactic variations while being at the same time compatible with the amount of available data.
Linguistic connections are then established by assessing pairwise distances based on the POS distributions.
Intriguingly, our analysis reveals definite clusters that correspond to well known language families and groups, with exceptions explained by distinct morphological typologies. Furthermore, we obtain a significant correlation between language similarity and geographic distance, which underscores the influence of spatial proximity on language kinships.
\end{abstract}


\maketitle

\section{\label{sec:intro}Introduction}

The number of languages in the world is estimated to be around 7,000~\cite{ethnologue}. This leads to a broad diversity at all linguistic levels: phonetic, morphosyntactic, semantic and pragmatic. The task to comprehend this immense variation is overwhelming. However, researchers have managed to pinpoint linguistic relationships that allow them to cluster languages in groups and families. Such classification was first based on comparative studies~\cite{hale2007historical,durie1996comparative} and now has increasingly been supported with quantitative approaches~\cite{gray2003language,gray2009language}.

A fertile approximation in historical linguistics describes languages as species in a phylogenetic tree that shows the evolutionary history from proto-languages to today's descents~\cite{greenhill2023language}. In this diachronic view, languages change through linguistic innovations that cause two languages with the same ancestor to become mutually unintelligible. There is also the complementary view, followed in this work, that searches for relations among languages in a given moment of time, in a synchronic manner~\cite{de2011course}, with the goal of quantifying the distance between languages using an appropriate metric~\cite{serva2008indo,holman2011automated}.

Language distances are encoded in matrices whose entries measure the similarity among certain linguistic features. The same method has been successfully applied in dialectometry, which aims at quantifying the regional differences among varieties of a given language~\cite{nerbonne2009data}. Quantitative measures of linguistic distances are useful not only for fundamental reasons but also in applied linguistics with the aim of analyzing the learning difficulties of minorities and immigrants~\cite{chiswick2005linguistic}. Another application of language distances is for languages in contact since languages that are more congruent to each other are more likely to coexist~\cite{mira2005interlinguistic}.
An assessment of linguistic similarities is therefore helpful for evidence-based policies and planning that seek to revitalize endangered languages~\cite{fernando2010model}.

Now, whilst most of the studies that compute distances either in synchronic linguistics or in dialectometry focus on orthographic, phonetic or lexical variations~\cite{nerbonne1997measuring,downey2008computational,heeringa2013lexical,donoso-sanchez-2017-dialectometric,gamallo2017language,eden2018measuring}, less attention has been paid to morphosyntactic features, other than a few exceptions~\cite{sanders2010statistical,longobardi2013toward,dunn2019global}. The latter are interesting because syntax is more robust to change than phonetics or semantics. Therefore, the resulting picture would show a larger time depth, and unique cases of accelerated change would hence stand out.

A particularly simple but elucidating approach to analyze syntactic variations is by means of parts of speech (POS). These denote word classes with well defined grammatical functions that share common morphological properties~\cite{manning1999foundations}. For instance, almost all languages distinguish between verbs and nouns, i.e., roughly between actions and entities. As a consequence, one can categorize words or lexical items as members of any of the proposed POS, typically around 15. This classification has its own limitations (e.g., certain languages do not distinguish between verbs and adjectives) but it has the advantage of simplicity while capturing at the same time a large amount of morphosyntactic information. The POS approach has been proven especially useful in natural language processing tasks. The reason is that POS reveal much not only about the syntactic category of a word but also about that of its neighboring words, due to semantic restrictions. For example, if a word is a noun it will most likely be surrounded by determiners and adjectives, forming a noun phrase. Therefore, we can gain insight about the phrasal structure of a language by examining POS sequences. This is precisely the main objective of this work: to model these sequences as stochastic processes, analyze their correlations and compute syntactic distances between languages using POS sequence distributions taken from a multilingual corpus.

Statistics of POS sequences, specifically the analysis of POS $r$-grams,
defined as sequences of $r$ successive POS, has served for diverse
research goals.
One may hypothesize that genres are characterized by different
syntactic structures, which would then allow for reliable genre classification. As demonstrated in Ref.~\cite{feldman2009part}, a careful study of POS trigram histograms provides a high-performance genre classifier. Strikingly, series of POS trigrams can be employed for building phylogenetic language trees solely from translations~\cite{rabinovich2017found}. The premise here is that syntactic features are retained in the translation process. 
Furthermore, assuming that POS tags can be predicted for historically close languages it is possible to train a language model
to measure proximity among languages~\cite{samohi2022using}.
However, these previous works set the POS block length in a somewhat heuristic manner. Below, we demonstrate using information-theoretic methods that trigrams suffice to account for the correlations present
in POS sequences. In other words, it is not necessary to consider
$r$-grams with $r\ge 4$ to gain more information, a result
that considerably simplifies analyses that involve POS series.

The mapping between a corpus and its corresponding POS series can be performed with human or automatic parsers. For the purposes of our investigation, we consider the Universal Dependencies library dataset~\cite{UD}, which includes manually annotated treebanks across many languages~\cite{de2021universal}. This dataset consists of texts and speech transcripts originated from various sources: news, online social media, legal documents, parliament speeches, literature, etc. 
The 17 universal parts of speech used in this particular dataset are grouped into three classes: open, which includes the tags adjective (ADJ), adverb (ADV), interjection (INTJ), noun (NOUN), proper noun (PROPN) and verb (VERB); closed, with the tags adposition (ADP), auxiliar (AUX), coordinating conjuntion (CCONJ), determiner (DET), numeral (NUM), particle (PART), pronoun (PRON) and subordinating conjunction (SCONJ); and others, which comprise punctuation (PUNCT), symbol (SYM) and other (X).

We then build a corpus of $67$ contemporary languages expressed by means of these tags. Since the number of possible POS $r$-grams grows exponentially with $r$, it is thus natural to ask what value of $r$ conveys the maximum information about a language. As aforementioned, we find that $r=3$ suffices to correctly characterize any of the studied languages. Then, we depict the connections between languages assessing the pairwise distance between POS trigram distributions. Interestingly, our found clusters can be identified with well known families and groups. Exceptions can be understood due to distinct linguistic typologies. This is natural since morphology constrains the possible POS combinations that can form and, consequently, this is reflected in the POS distributions and the distances calculated thereof.

Interestingly, we find a correlation between the obtained linguistic distance and the geographic distance spanned between locations assigned to each language. These are WALS locations, which generally correspond to the geographic coordinates associated to the centre of the region where the analyzed languages are spoken. However, for some languages, the regions where they are spoken are discontinuous. In such cases, the locations are placed within the larger region in which the language is spoken~\cite{wals}.
Despite the fact that the centres are obviously only approximate and that our calculation of the linguistic distances has its own limitations,
we clearly find that most of the syntactically close languages are also geographically close. This is expected since similar language usually lie in a continuum but there exist conspicuous exceptions, as we shall discuss below. 

\section{Methods}\label{sec:methods}
\subsection{Data}\label{sec:data}

The data utilized in this work is taken directly from the Universal Dependencies library~\cite{UD}, where each language is depicted with one or several corpora that are manually tagged. Thus, each word is classified  into one of the possible POS categories as stated above. These tags are deemed universal because they provide a common and consistent way to represent the grammatical categories of words across different languages. 
For example, the English sentence
\begin{align}
    &\text{Launching this way will hopefully avoid future disasters, giving more support towards}\nonumber \\ 
    &\text{NASA revisiting the stars.}
\end{align}
is converted into the POS sequence 
\begin{align}\label{eq:S}
  S&=\text{VERB, DET, NOUN, AUX, ADV, VERB, ADJ, NOUN,
  PUNCT, VERB, ADJ, }\nonumber \\ 
    &\text{NOUN, SCONJ, PROPN, VERB, DET, NOUN, PUNCT}  
\end{align}

Since we are interested in lexical classes that are either open or closed, we combine the 3 categories in the others class in a single tag. Hence, we will only consider $L=15$ distinct categories.
As a consequence, in alphabetical order, the possible tags are \\
$\lbrace z_i \rbrace_{i=0}^{14} = \lbrace \text{ADJ, ADP, ADV, AUX, CCONJ, DET, INTJ, NOUN, NUM, PART,}$  \\
$\text{PRON, PROPN, PUNCT, SCONJ, VERB}\rbrace$, where the categories in the others class are included in the tag PUNCT. 
We can then count the occurrences of each tag in $S$, excluding the period at the end of the sentence. For example, the number of times the tag $z_0=\text{ADJ}$ occurs in Eq.~\eqref{eq:S} is $2$. This way we gain access to unigram statistics. Similarly, we can group $S$ in overlapping blocks of $2$ consecutive tags and count the number of times each block is observed. For example, the block $(z_5,z_{7})=(\text{DET, NOUN})$ occurs in Eq.~\eqref{eq:S} $2$ times, thus opening the path to bigram statistics, etc. 

In general, the set of all $L^r$ possible $r$-grams, or blocks of size $r\geq 1$, is given by $\lbrace b_{j}^{(r)}: b_{(i_0,\ldots,i_{r-1})_L}^{(r)}=(z_{i_0},\ldots,z_{i_{r-1}}), \quad i_0,\ldots,i_{r-1} = 0,\ldots L-1 \rbrace^{L^{^{r}}-1}_{{{j=0}}}$, where $(i_0,\ldots,i_{r-1})_L$ is a base $L$ number.
Given the dataset of language $\mathcal{L}$, formed by $R$ tagged sentences, we count the number of appearances for each block of size $r$, with $1\leq r \leq  r_{\text{max}}$, with $r_{\text{max}}$ to be specified below.
First, we arrange each of the $R$ sequences in overlapping blocks of size $r$ and calculate the occurrences $\hat{n}_j^{(r)}$ of block $b^{(r)}_j$.
With this information we build for each value of $r$ the set of observations $\lbrace \hat{n}_j^{(r)} \rbrace_{j=0}^{L^{^r}-1}$ for every language $\mathcal{L}$.

$\mathcal{L}$ can be any of the $67$ contemporary European and Asian languages that fulfill the criterion of having datasets of at least $10$ thousand tokens
in the Universal Dependencies library. The comprehensive list
is included in App.~\ref{sec:appA}, along with their respective language family and group, as well as their corresponding morphological type (agglutinative, fusional, isolating). The latter is a useful information about word formation. We emphasize that these three types are just approximate categories since most of the languages have morphological traits of the three types with variable relevance~\cite{comrie1989language}.

Importantly, we select languages within a contiguous region (except for Afrikaans which is included since it belongs to the Indo-European family) because we shall later explore possible correlations between linguistic and geographic distances. Specifically, we focus on Eurasia since this is a single continent with both a rich linguistic diversity and abundant data availability.

Subsequently, we apply the previously discussed procedure of counting POS $r$-grams occurrences for every language within our dataset. Following an information-theoretic approach, we now argue in the next section that the $r$-gram probability distribution of each language for $r=3$ is sufficient to capture the correlations observed in POS sequences and, consequently, it can serve as a reliable basis for calculating distances between languages.

\subsection{Predictability gain and memory}\label{sec:memory}

The dynamics of many stochastic processes can be described by considering that the transition probabilities to future outcomes depend on previous states.
Consider a random variable $X$ with $L$ possible outcomes $z_0,\ldots,z_{{L-1}}$ and probability distribution $P(X) = \lbrace P(X=z_i), \quad i=0,\ldots,L-1 \rbrace$, where $P(X=z_i)$ is the probability that $X$ takes the value $z_i$. 
In our case, $X$ are the POS tags taking values specified above, and $P(X=z_i)$ is the probability for occurrence of tag $z_i$ in language $\mathcal{L}$.
Given $k+1$ repetitions of $X$, $X_0,\ldots,X_{k}$, the accuracy of  predicting the next outcome of the process, $X_{k+1}$, generally grows as the number of past states considered increases.
For example, the information contained in the stationary probability (or zeroth-order transition probability) $P(X_{k+1}=z_j)$ for outcome $z_j$ is always less or equally accurate than the information provided by the first-order transition probability $P(X_{k+1}=z_j|X_{k}=z_l)$.

The predictability gain $\mathcal{G}_u$ \cite{Crutchfield} quantifies the amount of information that one gains when performing predictions taking into account $u+1$ past states instead of $u$.
For completeness, we prove in App.~\ref{sec:appB} that, for homogeneous systems in which the transition probabilities are independent of time, the predictability gain takes the form
\begin{equation}
    \mathcal{G}_u = -(H_{u+2}-2H_{u+1}+H_u), \quad u \geq 0,
    \label{eq:pg2}
\end{equation}
where $H_r$ is the block Shannon entropy \cite{shannon} of size $r\geq 1$ ($H_0\equiv 0$). $H_r$  for $r\ge 1$ can be straightforwardly calculated from the joint probability distribution $P_r(X)$ of $r$ consecutive repetitions of the variable $X$,
\begin{equation}
    P_r(X) \equiv
    \lbrace P(X_0=z_{i_0},\ldots,X_{r-1}=z_{i_{r-1}}), \quad i_k = 0,\ldots,L-1 \rbrace,
    \label{eq:pr}
\end{equation}
as
\begin{align}
\begin{split}
H_r &= -\sum_{j=0}^{L^r-1} p(b_j^{(r)})\log (p(b_j^{(r)})),
\end{split}
\label{eq:block_entr}
\end{align}
where 
\begin{equation}
    p(b_{(i_0,\ldots,i_{r-1})_L}^{(r)}) = P(X_0=z_{i_0},\ldots,X_{r-1}=z_{i_{r-1}}), \quad 0\leq i_0,\ldots, i_{r-1} \leq L-1,
\end{equation}
is the probability for occurrence of block $b_{(i_0,\ldots,i_{r-1})_L}^{(r)}=(z_{i_0},\ldots,z_{i_{r-1}})$ with size $r$, and $\log$ is hereafter understood as $\log_2$. 

A stochastic process generated from consecutive repetitions of the variable $X$ has order or memory $m\geq 1$ if the transition probabilities satisfy
\begin{equation}
\begin{split}
P(X_{k+1}=z_j|X_{0}=z_v,\ldots,X_{k}=z_l) = 
P(X_{k+1}=z_j|X_{k-(m-1)}=z_o,\ldots,X_{k}=z_l).
\label{eq:transProbs}
\end{split}
\end{equation} 
These are usually referred to as $m$-th order Markovian processes \cite{raf85}. For the case $m=0$ the probabilities $P(X_{s}=z_j)$ are independent for all $s$. 

In Ref.~\cite{juan} it was shown that a system has memory $m$ if and only if $H_u$ is a linear function of $u$ for $u\geq m$. This result amounts to stating that a system has memory $m$ if and only if $\mathcal{G}_u=0$ for all $u\geq m$. This can be proven directly from Eq.~\eqref{eq:pg2}.
Hence, an analogous definition for the memory $m$ of a stochastic system follows:
\begin{equation}
    m = \text{min}(\lbrace \mu: \mathcal{G}_u=0 \text{, for all } u\geq \mu\rbrace).
    \label{eq:m}
\end{equation}
Therefore, $m$ is the minimum number of past states that we need to consider in order to achieve maximum predictability of the random process. We can use Eq.~\eqref{eq:m} as a benchmark to determine the memory of a system.
Moreover, if a stochastic process described with the random variable $X$ has memory $m$, knowing the joint probability distribution $P_{m+1}(X)$ of $m+1$ consecutive repetitions of $X$ is sufficient to capture all relevant information about the process since $P_{m+1}$ can be used to compute the probabilities of larger and smaller blocks. Hence, the $m+1$ block size is optimal for accurate predictions and understanding the process dynamics, without adding redundant information.

\subsection{Estimation from finite data}\label{sec:estimation}

Given the counts $\lbrace \hat{n}^{(r)} \rbrace\equiv \lbrace \hat{n}_j^{(r)} \rbrace_{j=0}^{L^{^r}-1}$ of language $\mathcal{L}$ obtained following the procedure outlined in Sec.~\ref{sec:data} for a value of $r$, we estimate the probability to observe block $b_j^{(r)}$ of $r$ consecutive POS tags in language $\mathcal{L}$ as 
\begin{equation}
    \hat{p}_{_\mathcal{L}}(b_j^{(r)}) = \dfrac{\hat{n}_j^{(r)}}{N^{(r)}},
    \label{eq:probs}
\end{equation}
where $N^{(r)} = \sum_{j=0}^{L^r-1}\hat{n}_j^{(r)}$. It is well known that Eq.~\eqref{eq:probs} is an unbiased estimator for the probability $p_{_\mathcal{L}}(b_j^{(r)})$.
Similarly, from our observations we can estimate the $(r-1)$th-order transition probabilities as
\begin{align}
    \begin{split}
\hat{p}_{_\mathcal{L}}(z_{i_{r}}|z_{i_1},\ldots,z_{i_{r-1}}) \equiv
\hat{P}(X_{r}=z_{i_{r}}|X_{1}=z_{i_1},\ldots,X_{r-1}=z_{i_{r-1}}) = \dfrac{\hat{n}^{(r)}_{(i_1,\ldots,i_{r-1},i_{r})_L}}{\sum\limits_{v=0}^{L-1}\hat{n}^{(r)}_{(i_1,\ldots,i_{r-1},v)_L}},
    \end{split}
    \label{eq:transprobs}
\end{align}
with $i_k = 0,\ldots,L-1$.

Our next step involves calculating the predictability gain in order to determine the memory of the POS sequences by means of Eq.~\eqref{eq:m}. Motivated by Eq.~\eqref{eq:pg2} we can perform this task by estimating the block entropies. 

Entropy estimation is a largely analyzed problem and, even though there
is not known unbiased estimator for the entropy \cite{paninski},
numerous useful estimators can be found in the literature \cite{contreras}.
Usually, all entropy estimators fail when the number of possible outcomes is larger than the available data. Considering that the number of possible blocks grows exponentially with increasing block size, there exists a certain $r_{\text{max}}$ for which entropy estimation is unreliable for $r > r_{\text{max}}$. The value of $r_{\text{max}}$ depends on the chosen entropy estimator, but it is usual to take $r_{\text{max}}\simeq \log(N^{(1)})/\log(L)$ \cite{juan}.

Hereafter we use the NSB entropy estimator \cite{nsb,nsb2}, which has shown to give good results for correlated sequences \cite{juan2}. Estimation of the block entropy of size $r$ using this method only requires knowledge of the data observations $\lbrace \hat{n}^{(r)} \rbrace$. The result of applying the estimator $\hat{H}$ to this set is denoted with $\hat{H}\left[\lbrace \hat{n}^{(r)} \rbrace \right]$, for $1 \leq r \leq r_{\text{max}}$.
Therefore, for finite data Eq.~\eqref{eq:pg2} becomes
\begin{equation}
\hat{\mathcal{G}}_u[\lbrace \hat{n} \rbrace] = -\left(\hat{H}\left[\lbrace \hat{n}^{(u+2)} \rbrace \right]-2\hat{H}\left[\lbrace \hat{n}^{(u+1)} \rbrace \right]+\hat{H}\left[\lbrace \hat{n}^{(u)} \rbrace \right] \right), \quad 1\leq u \leq r_{\text{max}}-2,
    \label{eq:pg_est}
\end{equation}
and
\begin{equation}
    \hat{\mathcal{G}}_0[\lbrace \hat{n} \rbrace] = -\left(\hat{H}\left[\lbrace \hat{n}^{(2)} \rbrace \right]-2\hat{H}\left[\lbrace \hat{n}^{(1)} \rbrace \right] \right).
    \label{eq:pg_est0}
\end{equation}

Due to the limitations of entropy estimation, the condition imposed in Eq.~\eqref{eq:m} to determine the value of $m$ is too strict. Rather than attempting to ascertain the minimum block size at which the predictability gain is $0$, a more sensible approach is to compare the values of $\hat{\mathcal{G}}_u[\lbrace \hat{n} \rbrace]$ with those obtained from a scenario where we know that the estimated predictability gain should be $0$.

\subsection{POS trigrams}\label{sec:motivation}

In this section we provide evidence supporting our claim that in order to capture the correlations in POS sequences it is enough to consider their trigram probability distribution. We show this in two ways: in Sec.~\ref{sec:pred_gain}, calculating the predictability gain, and in Sec.~\ref{sec:markov_models}, analyzing the accuracy in language detection as the order of Markovian models increases.

\subsubsection{Predictability gain}\label{sec:pred_gain}

Our first objective is to quantify the information gained in POS sequences of various languages when performing predictions considering $(u+1)$th-order transition probabilities rather than $u$th-order.

\begin{figure}[t]
\sidesubfloat[]{\includegraphics[width=0.43\columnwidth]{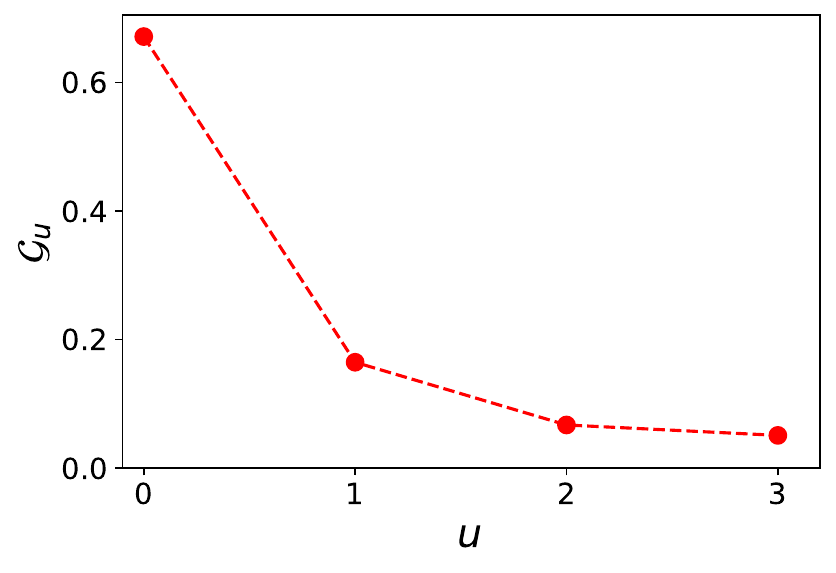}}
 \quad
\sidesubfloat[]{\includegraphics[width=0.43\columnwidth]{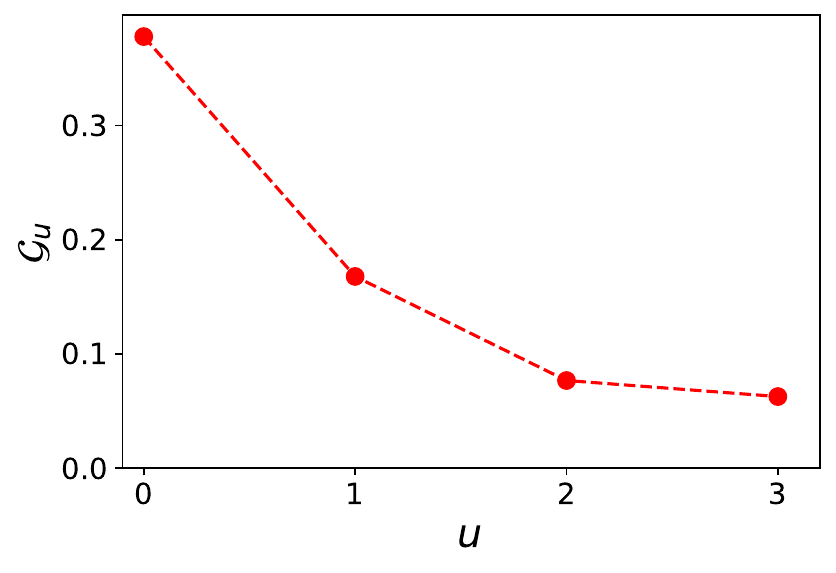}} \\
\sidesubfloat[]{\includegraphics[width=0.43\columnwidth]{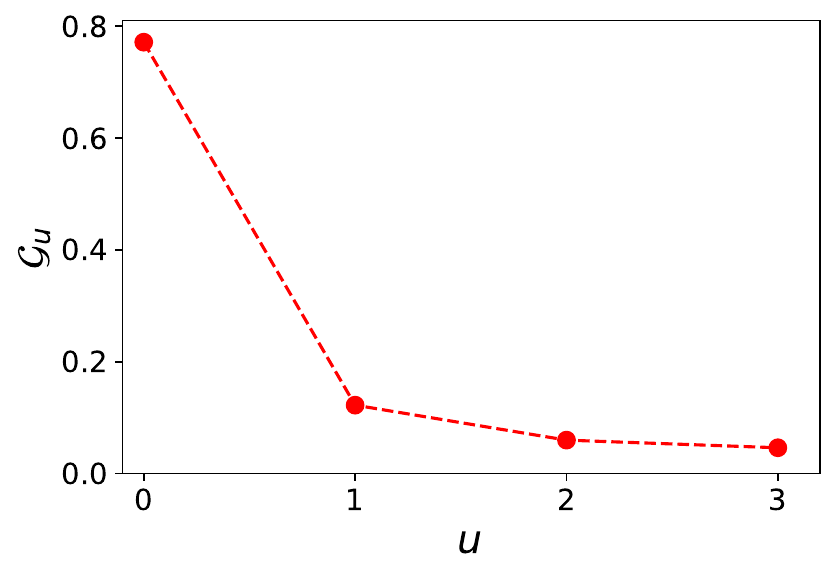}} 
\quad
\sidesubfloat[]{\includegraphics[width=0.43\columnwidth]{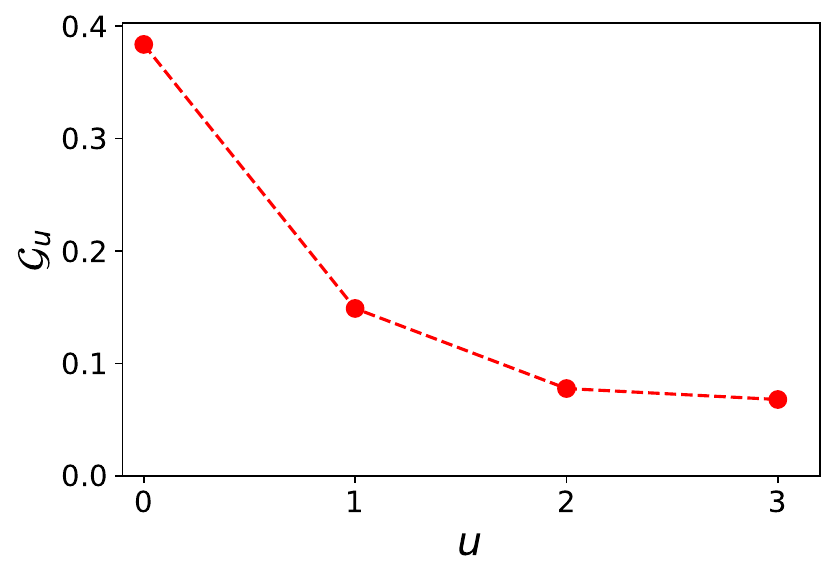}}
 \caption{Estimated predictability gain when considering $(u+1)$th-order instead of $u$th-order transition probabilities in POS sequences of (a) German, (b) Icelandic, (c) Portuguese and (d) Czech, as extracted from the Universal Dependencies library.}
 \label{fig:pg}
\end{figure}

In Fig.~\ref{fig:pg} we show $\hat{\mathcal{G}}_u$ for
two representative languages of the same group, namely, German (panel a) and Icelandic (panel b) from the Germanic group, and two representative languages of different groups
that also differ from the previous one, namely, Portuguese (panel c)
from the Romance group and Czech (panel d) from the Slavic group.
This way we can test intragroup and intergroup variations, if they exist. Further, these three groups constitute the most extensive clusters, characterized by a wealth of available data, which make the results more trustworthy.
We set $r_{\text{max}}=5$ for the four languages; hence we ensure that the estimation of their predictabilities is reliable up to $u=3$.

\floatsetup[figure]{style=plain,subcapbesideposition=top}

The values of $\hat{\mathcal{G}}_u$ at $u=0$ for the four languages demonstrate a significant predictability gain when transitioning from zeroth-order predictions to first-order predictions, suggesting that the POS state of the sequence at a given step $k$ is highly informative about the next POS outcome at step $k+1$. We observe that the values of $\hat{\mathcal{G}}_0$ significantly vary for each language, even for German and Icelandic which belong to the same group. Then, the  $\hat{\mathcal{G}}_u$ values at $u=1$ indicate that the information provided by the preceding POS outcome at step $k-1$ is also substantial in terms of predictability. In contrast, the predictability gain at $u=2$ drops to $\hat{\mathcal{G}}_2 \simeq \dfrac{\hat{\mathcal{G}}_1}{2}$ and remains relatively constant at $u=3$.
We obtain this decreasing pattern in the curves of $\mathcal{\hat{G}}_u$ as a function of $u$ for all languages considered in this work. Even though this does not prove that POS sequences have memory $2$ (see App.~\ref{sec:appC} for further details), our results indeed show that considering transition probabilities beyond order $2$ does not yield substantially more information about the correlations present in our POS sequences. 

\subsubsection{Markov models}\label{sec:markov_models}

We now provide more evidence that POS trigrams (or equivalently, POS sequences of memory $2$) describe most of the statistical information
contained in the represented languages. 

From the corpus of language $\mathcal{L}$ we extract a tagged sentence $S=x_{_1},\ldots,x_{_N}$ of length $N$ and compute the probability $P(S|\mathcal{L}')$ of observing $S$ in a certain language $\mathcal{L}'$.
This probability can be computed either by considering the estimated stationary distribution of language $\mathcal{L}'$,
\begin{equation}
\hat{P}^{(0)}(S|\mathcal{L}')=\prod_{k=1}^N \hat{p}_{_{\mathcal{L}'}}(x_k),    
\end{equation}
or by incorporating $u$th-order transition probabilities for $u\geq 1$ ($u$-order Markov model).
For example, the probability of observing the sequence $S$ given the language $\mathcal{L}'$ considering first-order transition probabilities
reads
\begin{equation}
\hat{P}^{(1)}(S|\mathcal{L}') = \hat{p}_{_{\mathcal{L}'}}(x_1)\prod_{k=1}^{N-1} \hat{p}_{_{\mathcal{L}'}}(x_{k+1}|x_{k}).
\end{equation}
Similarly, we can compute $\hat{P}^{(u)}(S|\mathcal{L}')$ for $u>1$.
This procedure is repeated for various languages, and the language that yields the highest probability is assigned to the one from which $S$ is generated. Ideally, the obtained language would correspond to $\mathcal{L}$ for all values of $u$.

For each of the four previously considered languages (German, Icelandic, Portuguese, and Czech) we randomly select from their respective corpora $K$ sentences $S_1,\ldots,S_K$, each  comprising 5 to 20 word tokens. Subsequently, for every sentence we compute $\hat{P}^{(u)}(S_l|\mathcal{L}')$ for the four languages and for values of $u$ ranging from $0$ to $3$.
The accuracy $A_u$ of correctly identifying the language associated with each case is then assessed as the fraction of correct sentence classifications:
\begin{equation}
A_u(\mathcal{L}) = \dfrac{\#\lbrace l: \hat{P}^{(u)}(S_l|\mathcal{L}) > \hat{P}^{(u)}(S_l|\mathcal{L}') \text{ for all } \mathcal{L}'\neq \mathcal{L} \rbrace_{l=1}^{^K}}{K}.
\end{equation}

\begin{figure}[t]
\includegraphics[width=.7\textwidth]{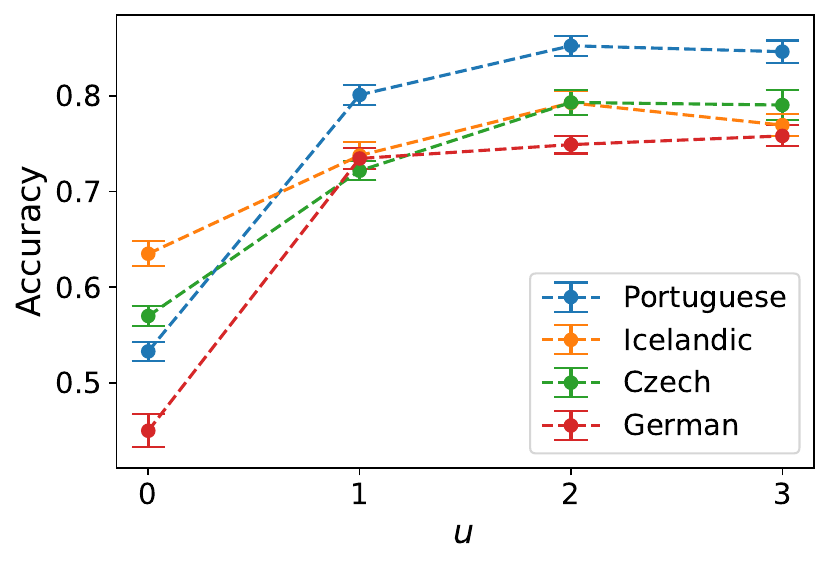}
 \caption{Accuracy in language identification, determined by computing and comparing the probabilities of observing a given tagged sentence within each considered language, based on an $u$th-order Markov model.}
 \label{fig:markov_model}
\end{figure}

We compute the estimated stationary and transition probabilities, as specified in Eq.~\eqref{eq:probs} and \eqref{eq:transprobs} respectively, without considering the influence of the $K$ sentences utilized for testing purposes.
We then repeat this procedure $10$ times to find the mean and standard deviation of the accuracy for each language, as a function of $u$. We consider $K=1000$. The results are displayed in Fig.~\ref{fig:markov_model}.

We observe that the accuracy for all languages significantly increases when first-order transition probabilities are considered ($u=1$). Subsequently, there is a slight increase at $u=2$, and afterwards the accuracy remains relatively constant for $u=3$. 
These findings agree with the analysis performed for the predictability gain, overall indicating that considering memory values higher than $2$ does not provide significant additional information. Conversely, choosing a low memory value proves advantageous as it enhances the probability estimation accuracy. Consequently, this enables us to incorporate a broader set of languages into our analysis.

\subsection{Language distances}
\label{sec:distances}

As explained in Sec.~\ref{sec:memory}, if a stochastic process generated from repetitions of the variable $X$ has memory $m$, the probability distribution $P_{m+1}(X)$ is adequate for capturing all relevant information about the process. 
We previously showed that sequences of POS tags can be modeled as processes with memory $2$ with high accuracy. Therefore, we can define a distance metric between languages $\mathcal{L}$ and $\mathcal{L}'$ from the statistical distance between their corresponding trigram distributions. To this end,
we consider the Jensen-Shannon (JS) distance \cite{jsdistance}.
Defining $\lbrace b_j\rbrace_{{j=0}}^{L^{^{3}}-1}$ as the set of all possible POS trigrams, the JS distance between languages $\mathcal{L}$ and $\mathcal{L}'$ is determined by
\begin{align}
\begin{split}
    d_{_{JS}}(\mathcal{L},\mathcal{L}') &= \\
    &\sqrt{\dfrac{1}{2}\sum_{j=0}^{L^{^{3}}-1}\left( p_{_\mathcal{L}}(b_j)\log\left(\dfrac{2p_{_\mathcal{L}}(b_j)}{p_{_\mathcal{L}}(b_j)+p_{_{\mathcal{L}'}}(b_j)} \right) + p_{_{\mathcal{L}'}}(b_j)\log\left(\dfrac{2p_{_{\mathcal{L}'}}(b_j)}{p_{_\mathcal{L}}(b_j)+p_{_{\mathcal{L}'}}(b_j)}\right) \right)}\,.
    \label{eq:js}
    \end{split}
\end{align}
It is worth noting that the $d_{_{JS}}$ measure satisfies all the essential properties expected for a metric and that $d_{_{JS}}$ ranges between
0 and 1. 

\floatsetup[figure]{style=plain,subcapbesideposition=top}
\begin{figure}[t]
\sidesubfloat[]{\includegraphics[width=0.44\columnwidth]{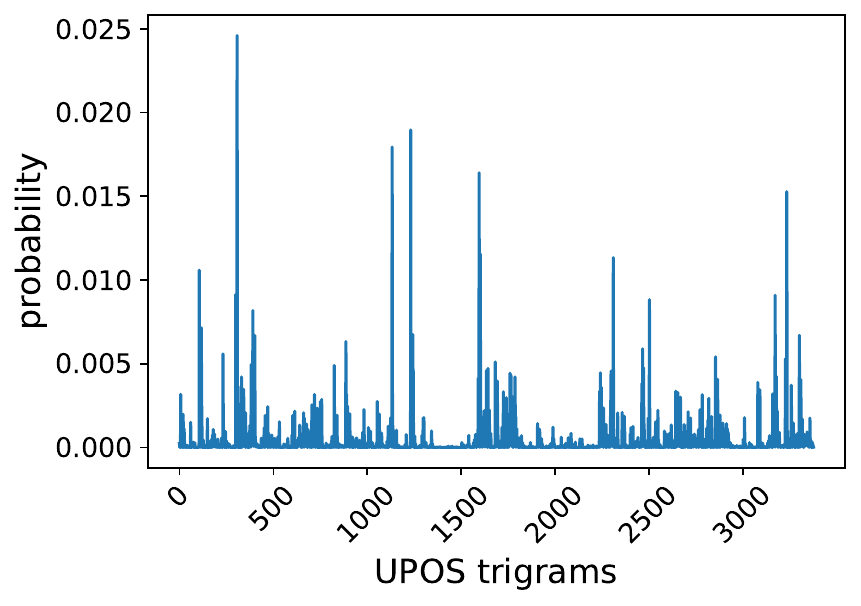}}\quad
\sidesubfloat[]{\includegraphics[width=0.44\columnwidth]{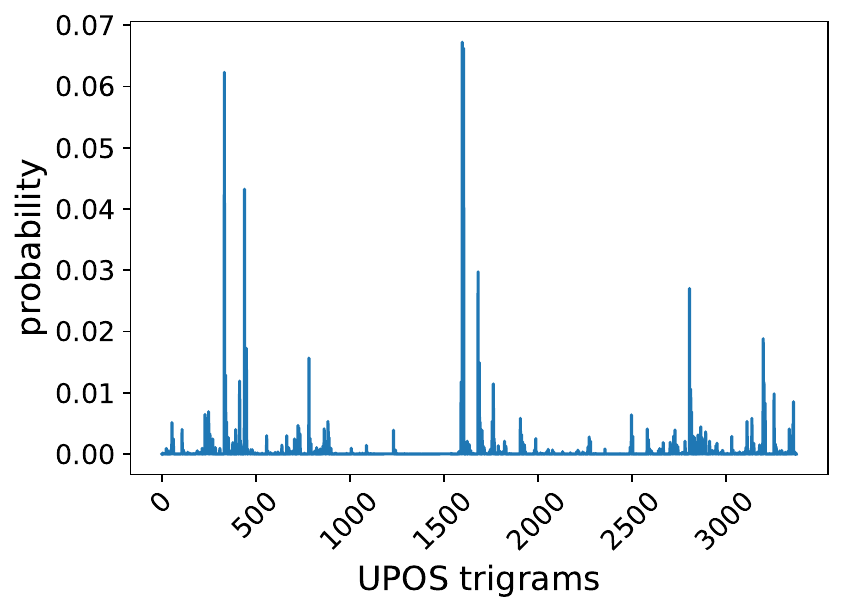}} 
 \caption{Probability distribution of POS trigrams for (a) English and (b) Japanese.}
 \label{fig:trigrams}
\end{figure}

We estimate the JS distance by replacing the exact probabilities $p(b_j)$ in Eq.~\eqref{eq:js} with the maximum likelihood estimators given by Eq.~\eqref{eq:probs}. As an illustration, we present in Fig.~\ref{fig:trigrams} the trigram probability distributions for English (panel a) and Japanese (panel b). The numbers that designate the trigrams correspond to their index. For example, trigram~0 refers to the block $b_0=(z_0,z_0,z_0) = (\text{ADJ, ADJ, ADJ})$; trigram~1 is $b_1=(z_0,z_0,z_1) = (\text{ADJ, ADJ, ADP})$; and so on until the last trigram 3374, which corresponds to the block $b_{3374}=(z_{14},z_{14},z_{14}) = (\text{VERB, VERB, VERB})$. 

From a simple inspection of Fig.~\ref{fig:trigrams} it is clear that the trigram probability distributions of English ($\mathcal{E}$) and Japanese ($\mathcal{J}$) show substantial differences. For example, in English the three most probable trigrams are $b_{307}=\text{(ADP, DET, NOUN)}$, $b_{1231}=\text{(DET, NOUN, ADP)}$ and $b_{1132}=\text{(DET, ADJ, NOUN)}$ whereas for Japanese we have $b_{1597}=\text{(NOUN, ADP, NOUN)}$, $b_{1604}=\text{(NOUN, ADP, VERB)}$ and $b_{331}=\text{(ADP, NOUN, ADP)}$. These trigrams differ because determiners are generally absent from Japanese, unlike English, and adpositions follow the Japanese nouns whereas in English adpositions can also appear before the noun. These differences between the two languages can be quantified using Eq.~\eqref{eq:js}. We find $d_{_{JS}}(\mathcal{E},\mathcal{J})=0.79$, which is a high value
due to the strong morphosyntactic differences between Japanese and English.

Following a similar methodology, for each language pair within our dataset we compute their corresponding JS distances using Eqs.~\eqref{eq:js}, yielding a distance matrix of dimension $67\times 67$, on which our subsequent clustering analysis is based. This methodology considers all available data for each language. An alternative approach is presented in App.~\ref{sec:appD}, where we calculate distances among text samples extracted from the same language group. We now present the results obtained with the JS distance.
Importantly, we also obtain similar results employing a different metric (the Hellinger distance, see App.~\ref{sec:appHellinger}), which reinforces the validity of our findings.

\section{Results}

We first discuss the general results obtained from calculating distances between languages based on their POS distributions. Then, we inquire a possible correlation between linguistic and physical (i. e., geographical) distances.

\subsection{Language distances and cluster analysis}

\subsubsection{Distance matrix}
\label{sec:dist_matrix}

\begin{figure}[ht!]
\includegraphics[width=1\textwidth]{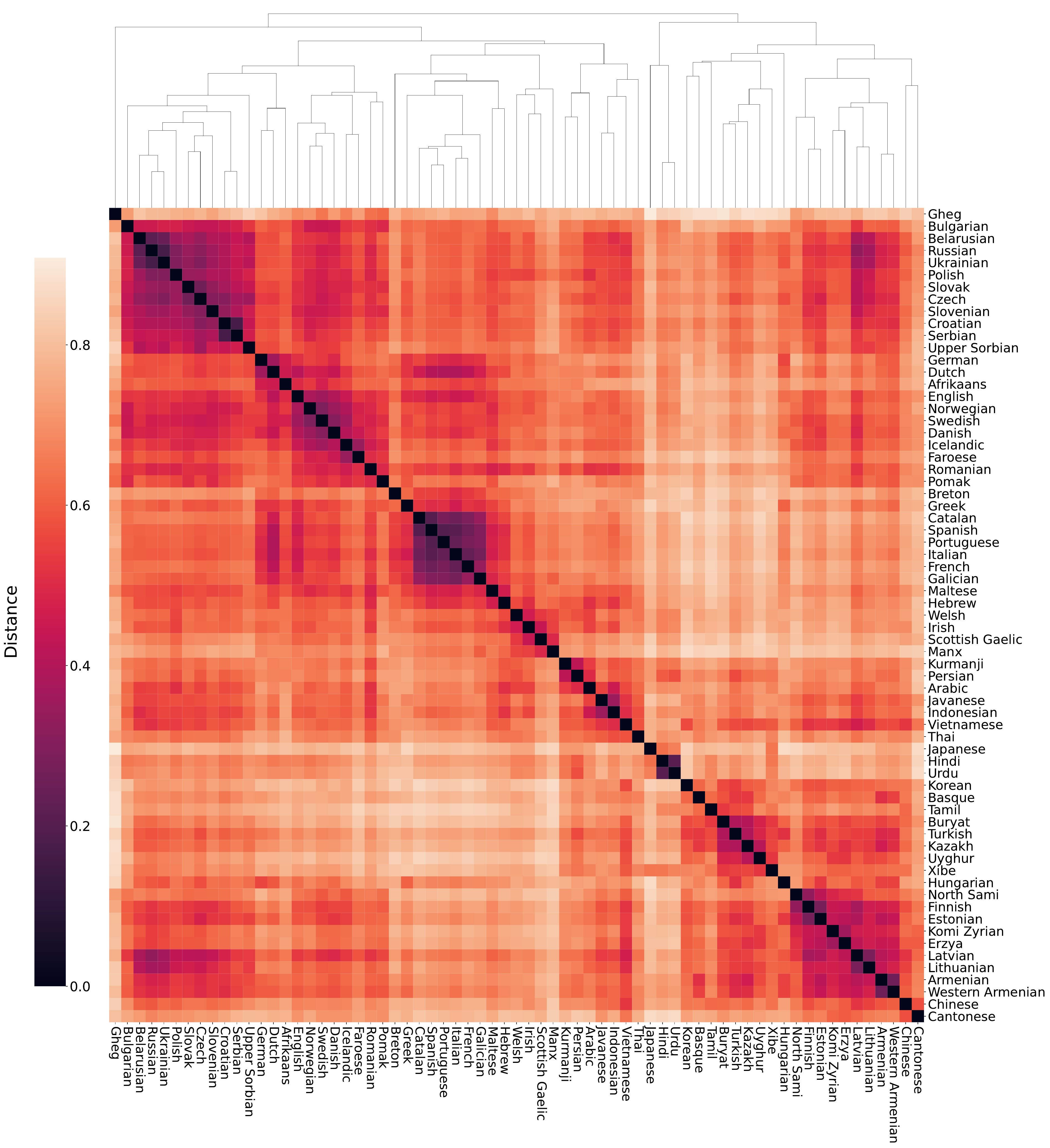}
 \caption{Heatmap visualization of the Jensen-Shannon distance matrix, calculated from POS trigram distributions. Rows and columns are organized based on hierarchical clustering. The colour spectrum in the heatmap illustrates data matrix values.}
 \label{fig:colormap_JS}
\end{figure}

The distance matrix generated from the data with the aid of Eq.~\eqref{eq:js} can be better visualized through a clustermap (see Fig.~\ref{fig:colormap_JS}). This representation employs hierarchical clustering~\cite{Nielsen2016} and heatmap visualization. We use the complete linkage method~\cite{complete} for clustering, organizing rows and columns based on similarity, depicted as dendrograms in the same figure. The heatmap, which represents distance values by a color spectrum as indicated in Fig.~\ref{fig:colormap_JS}, enables a comprehensive exploration of our data relationships and structure.

First, since darker colours indicate shorter distance (i.e, greater morphosyntactic similarity) one can clearly distinguish large-scale cluster formation. In the vicinity of the left upper corner we observe the most extensive cluster, corresponding to Slavic languages. Within this language group, we discern smaller clusters, the most clear being among Belarusian, Russian and Ukrainian (the East branch of the Slavic family), as well as between Serbian and Croatian (the South branch). Interestingly, our results point to a close relationship between this group and two members of the Baltic group (Latvian and Lithuanian). This suggests that spatial proximity might be correlated with our POS distances, as we discuss below.

Germanic languages exhibit a more dispersed pattern, manifested itself in two distinct clusters. The first cluster encompasses Afrikaans, Dutch, and German (the West branch), while the second is primarily composed of North Germanic languages along with English, which highlights its mixed character via a close connection with Romance languages. The latter, in turn, exhibit the most compact clustering, as evidenced by their tight proximity to one another in the heatmap. An exception is Romanian, which falls outside this group, probably because it is the only major Romance language with noun declension and article enclitics. Also, its spatial distance from the West may play a role. This interpretration is supported by the fact that Romanian is clustered with Pomak, a Bulgarian variety spoken in the geographically close region of Thrace.

We next find a small cluster that encompasses Maltese and Hebrew. Surprisingly, Arabic, classified as a Semitic language as well, appears closer to Austronesian languages, such as Indonesian and Javanese, as well as Persian. The reason lies in the different typology of Arabic and its
strong dialect diversity. Further visible clusters correspond to Celtic languages, Hindi and Urdu (both belonging to the Indic language group), the Turkic family (Turkish, Kazakh and Uyghur), adjacent to Buryat (a Mongolic language), followed by a larger cluster primarily comprising Uralic, Tungusic, and Sino-Tibetan families. This is the most diverse cluster
partly because the Universal Dependencies library
contains data of a few languages only as compared with the previous
families. However, there exist interesting connections that we
explore in the next sections.

\subsubsection{Language tree and $k$-medoids clustering}
\label{sec:clusters}

In order to gain further insight on the relations among the distinct languages, we perform a $k$-medoids clustering upon the distance matrix. Hence, we utilize the Partition Around Medoids (PAM) algorithm \cite{kaufman2009finding}, for which
the optimal number of clusters is determined through a silhouette analysis~\cite{ROUSSEEUW198753}. The corresponding figure and a list of languages constituting each cluster are presented in App.~\ref{sec:appF}. For the moment, we will make use of these results to build a single picture of the main interlinguistic connections.

\begin{figure}[ht!]
\includegraphics[width=1\textwidth]{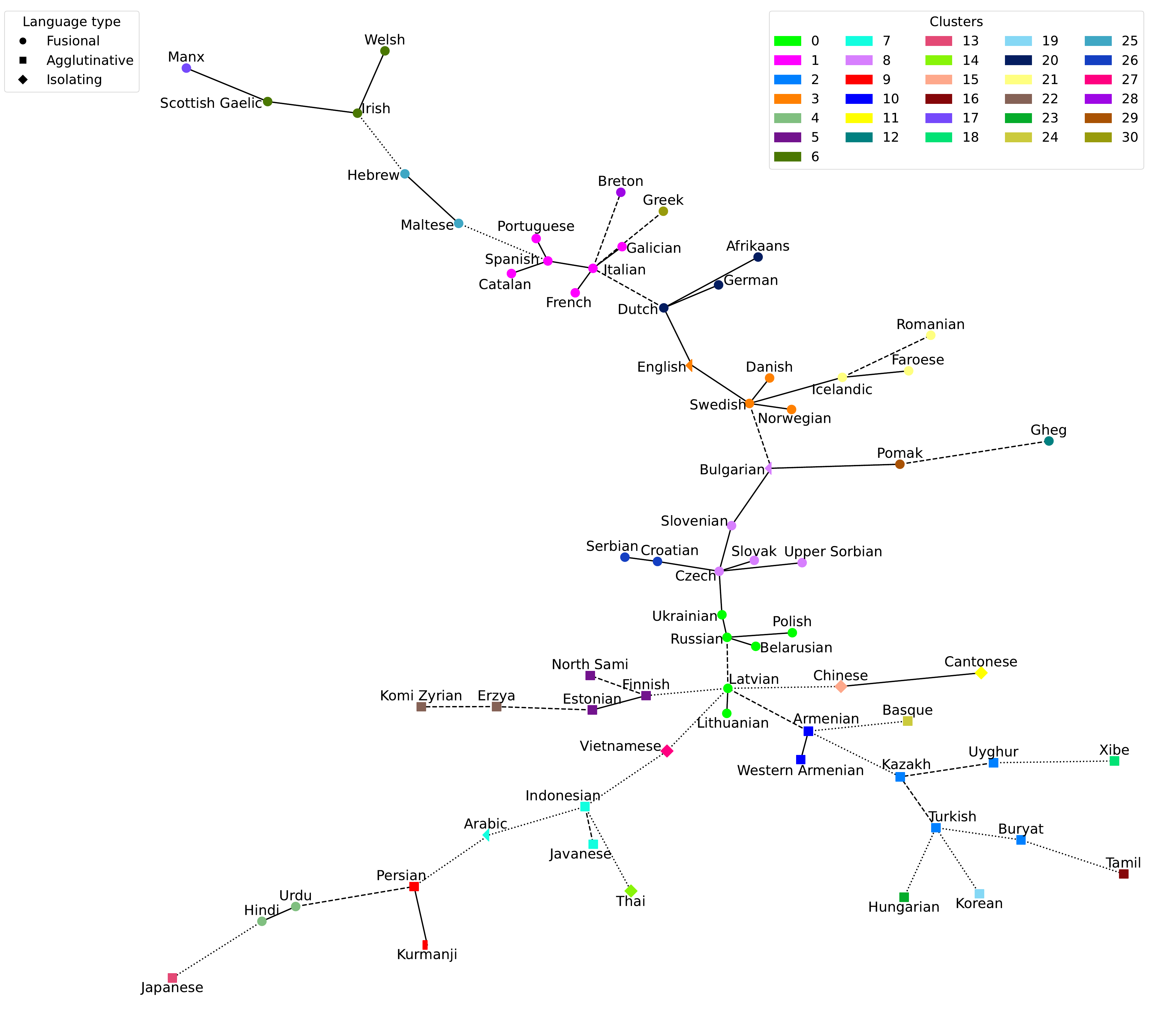}
 \caption{Minimum spanning tree generated from the Jensen-Shannon distance matrix, with node colors representing clusters identified through $k$-medoids analysis. The shape of the nodes represents the language typology. Full lines are assigned to links between languages belonging to the same group; dashed lines for languages of the same family but different group; finally, dotted lines connect languages from distinct families.}
 \label{fig:mst_js}
\end{figure}

First, we construct a network whose nodes represent languages whereas the edges' weight are given by their pairwise distances. Only the edges that generate a minimum spanning tree are considered as part of our analysis. The minimum spanning tree is a structure that connects all the data points in our dataset results with the minimum possible total edge weight~\cite{gower1969minimum}. We depict the tree in Fig.~\ref{fig:mst_js}, employing the Kamada-Kawai layout~\cite{kamada1989algorithm}. This is a force-directed layout, implying  that, for weighted graphs as in our case, the length of the edges tends to be larger for heavier weights. 
This visualization provides an intuitive representation of the relationships encoded in the minimum spanning tree, offering insights into the overall structure and connectivity patterns, not only among closely related languages, but also between distinct language families.
To enhance our interpretability, we assign consistent colors to items within the same cluster, obtained from the $k$-medoids analysis mentioned earlier. Nodes from the same language group are linked with a full line; nodes from the same family but different group are linked with a dashed line; finally, nodes belonging to distinct families are connected with dotted lines.
The shape of the nodes represents the language type: circles are assigned to fusional languages, squares to the agglutinative type and diamonds to isolating languages. The languages plotted with half nodes are both isolating and another type. For example, English is plotted with a half diamond node because it is considered to be a fusional-isolating language~\cite{haselow2011typological}. 

Interestingly, the same cluster formations observed with our previous analysis of Fig.~\ref{fig:colormap_JS} naturally emerge in the constructed tree, as well as connections between language families hidden within the distance matrix, which are easier to identify in this simpler visualization. We can find most of the Celtic languages (clusters $6$, $28$) close to a few Semitic languages (cluster $25$), specially Hebrew, whereas Maltese can be found in close proximity to Romance languages, which are once again forming a compact cluster ($1$), with the exception of Romanian which is clustered with Icelandic and Faroese (cluster $21$).
The Germanic group is subdivided into three groups (clusters $3$, $20$ and $21$) that are relatively close to one another, and similarly with the Slavic group (clusters $0$, $8$, $26$, $30$).
The Baltic languages are clustered together with the Slavic sub-group mostly consisting in East Slavic languages (cluster $0$). The proximity of these two groups was also observed in Fig.~\ref{fig:colormap_JS}.

The Baltic languages are connected with the Uralic languages (clusters $5$, $22$) and with the Armenian family (cluster $10$). An exception is Hungarian, considered an Uralic language, but here appearing separately (cluster $23$) near the Turkic languages, which are grouped together as expected, alongside Buryat (cluster $2$). Around this cluster several isolated languages such as Korean (cluster $19$), Tamil (cluster $16$) and Xibe (cluster $18$) can be found. 

Arabic is clustered with Austronesian languages (cluster $7$), in close proximity to the Iranian cluster (cluster $9$). The Indic languages (cluster $4$) are noticeably distant from the rest, and even further is Japanese (cluster $13$). The remaining clusters are formed by only one language, namely: Gheg ($12$), Thai ($14$), Basque ($24$), Greek ($30$), Cantonese ($11$) and Chinese ($15$). With the exception on the latter two, which belong to the Sino-Tibetan family, all others are the sole representatives of their respective language groups being considered in this analysis.

It is important to note that, even though this form of visualization can be helpful to recognize clusters and connection among languages, some of these connections are spurious and mainly arise from the inherent nature of the minimum spanning tree, which is designed to avoid leaving isolated nodes.
Nevertheless, it is remarkable that most of the connections observed in Fig.~\ref{fig:mst_js}, represented both by the cluster identification of the nodes and the edges linking them, are between languages belonging to the same group. Moreover, most of the languages displayed in the upper part of the figure belong to the Indo-European family, and they all have fusional features. On the other hand, the lower part of the chart mostly consists of agglutinative and isolating languages. 

Less known connections, such as among Hebrew and Celtic languages, or between Basque and Armenian are also displayed in the tree. These similarities have been already pointed out in Refs.~\cite{gensler1993typological} and \cite{tamrazian1994syntax} respectively, with the proper observation that this does not imply a common origin of the languages. A similar case is that of Altaic languages~\cite{AltaicLanguages}, a putative family relationship between Turkic, Mongolic, Tungusic, Korean and Japanese languages, a subject characterized by ongoing debate within linguistic research~\cite{janhunen2023unity}. In Fig.~\ref{fig:mst_js} this group is represented in the lower part of the tree by the connections among Turkish, Kazakh, Uyghur, Xibe, Buryat and Korean. Despite the controversy surrounding the existence of a genetic relationship, it is generally acknowledged that there exist linguistic similarities among these languages, as evidenced by our analysis.

Romanian language, which is considered to be a Romance language, can be found in Figs.~\ref{fig:colormap_JS} and \ref{fig:mst_js} closer to Germanic languages than to other Romance languages. Interestingly, this fact has been observed in Ref.~\cite{samohi2022using}, also using Universal Dependencies data but considering a machine learning approach to calculate linguistic distances.
In Ref.~\cite{rabinovich2017found}, a similar cluster analysis based on the analysis of POS trigrams of translations locates Romanian language away from the Romance group. 

Another interesting case is that of Arabic, a Semitic language which, in the dendrogram in Fig.~\ref{fig:colormap_JS} is clustered alongside Persian and Kurmanji, both Iranian languages. In the tree presented in Fig.~\ref{fig:mst_js}, Arabic and Persian are linked but the former is clustered by the $k$-medoids algorithm with Indonesian and Javanese, which belong to the Austronesian family. 
In this figure, Arabic appears in the lowest part of the tree, very distant from the other Semitic languages, Hebrew and Maltese. However, this is misleading because their linguistic distance in Fig.~\ref{fig:colormap_JS} is not high, specially between Arabic and Hebrew.
Further, the linguistic proximity observed between Arabic and Persian is influenced by their geographic closeness.

Overall, our cluster results are consistent with well established
families and linguistic groups. We indeed observe departures that could be
attributed to methodology inaccuracies. However, another interpretation
is possible. Very recently, a family tree based on the phylogenetic signal of syntactic data has been inferred~\cite{hartmann2024strength},
pointing to salient deviations
with respect to the trees derived from the comparative method,
which typically did not take into account syntactic data.
Therefore, language groups formed upon syntactic analyses
need not fully agree with those groups that emerge
from phonetic or lexical similarities.

In App.~\ref{sec:appG}, we present the results of a similar analysis to the one conducted in this section using tetragrams instead of trigrams. The findings from both methods are consistent, providing strong evidence for our claim that the probability distribution of POS trigrams is sufficient to capture syntactic information in the analyzed languages.

Strikingly, our results also suggest correlations between
syntactic distances and spatial proximity. In the following section,
we explore the connection between these two variables, which can account for
the relationship observed in Fig.~\ref{fig:mst_js} between, e.g.,
Semitic and Romance languages, Indonesian with Vietnamese or Uralic with
Slavic languages.

\subsection{Relation between linguistic and geographic distances}

We calculate the geographic (geodesic) distances between all language pairs by assigning an spatial coordinate to each language considered. This geolocation information is obtained from the World Atlas of Language Structures Online (WALS)~\cite{wals}. For this analysis we exclude Afrikaans since it is geographically isolated from the rest of the languages considered.

\floatsetup[figure]{style=plain,subcapbesideposition=top}
\begin{figure}[t]
\sidesubfloat[]{\includegraphics[width=0.44\columnwidth]{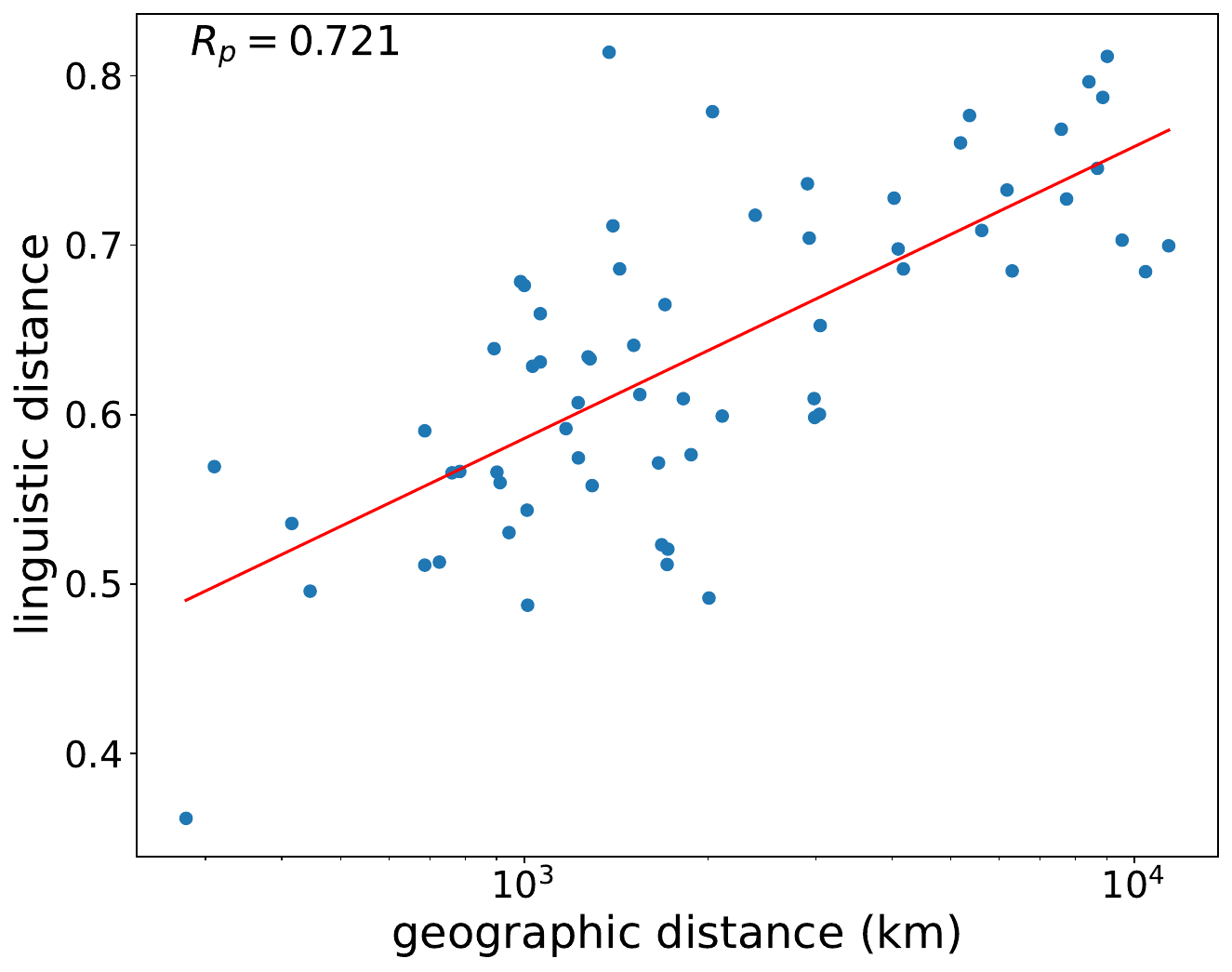}}
 \quad
\sidesubfloat[]{\includegraphics[width=0.44\columnwidth]{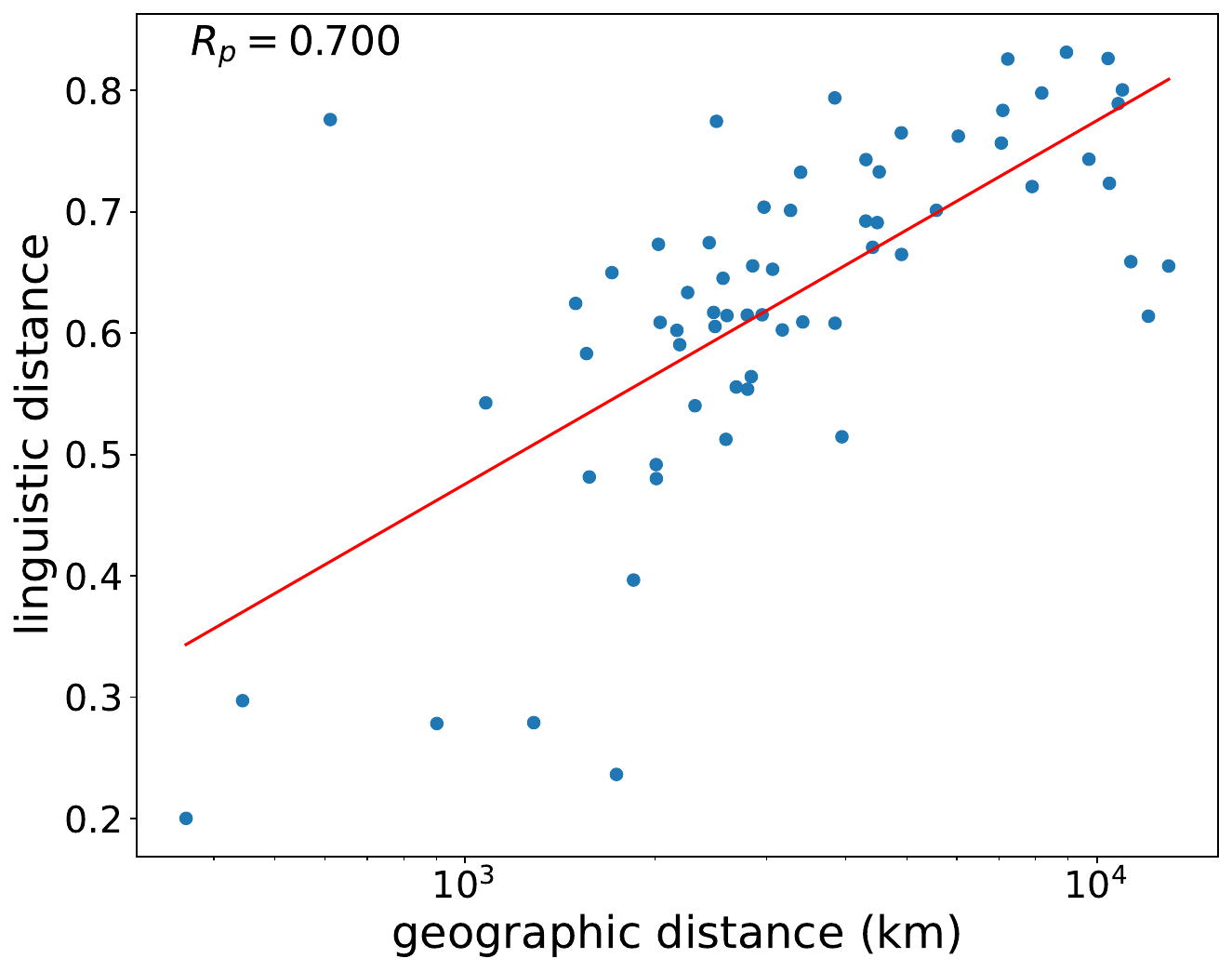}} \\
\sidesubfloat[]{\includegraphics[width=0.44\columnwidth]{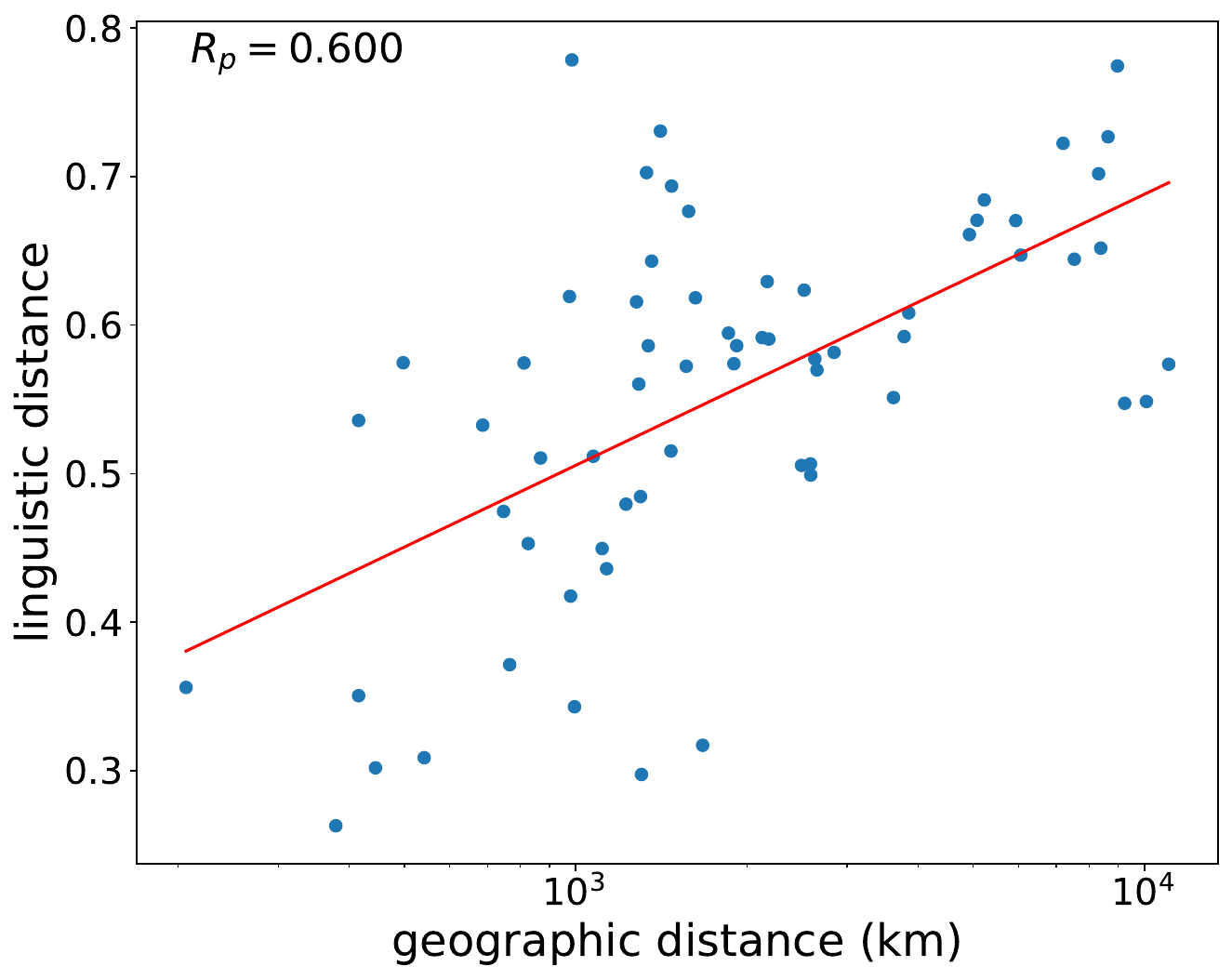}}
 \quad
\sidesubfloat[]{\includegraphics[width=0.44\columnwidth]{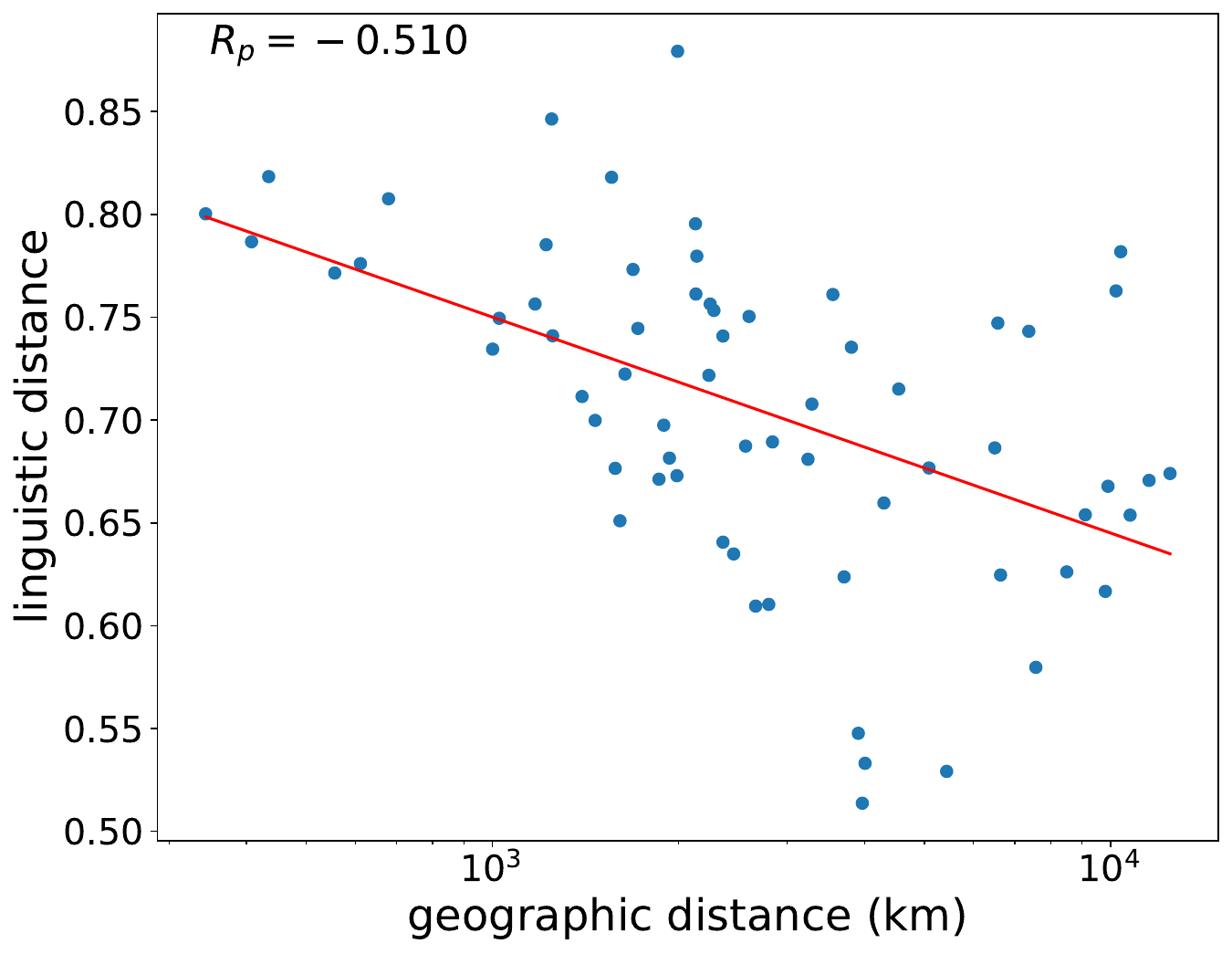}}
 \caption{Linguistic versus geographic distances between a selected language and all other languages considered: (a) German, (b) Portuguese, (c) Czech and (d) Basque. $R_p$ denotes the Pearson correlation coefficient.}
 \label{fig:ling-geo}
\end{figure}

First, for a single language we compute its linguistic and geographic distances with the rest of considered languages. The plots obtained for German (panel a), Portuguese (panel b), Czech (panel c) and Basque (panel d) are presented in Fig.~\ref{fig:ling-geo}, applying logarithmic scale to the geographic distances. The obtained Pearson correlation coefficients ($R_p=0.721$, $0.700$, $0.600$, and $-0.510$, respectively) point to a 
logarithmic relation between the two distances (all p-values are $<0.001$).
Surprisingly, out of the $66$ languages analyzed, only Basque [Fig.~\ref{fig:ling-geo}(d)] shows a significant negative correlation between linguistic and geographic distances. This is because
Basque is categorized as an agglutinative language and while the majority of Western European languages lean towards fusional characteristics, agglutinative languages are predominantly found in Eastern Europe and Asia. Consequently, Basque shares more linguistic similarities with languages spoken at greater distances than with those geographically closer.

\begin{figure}[t]
\includegraphics[width=.6\textwidth]{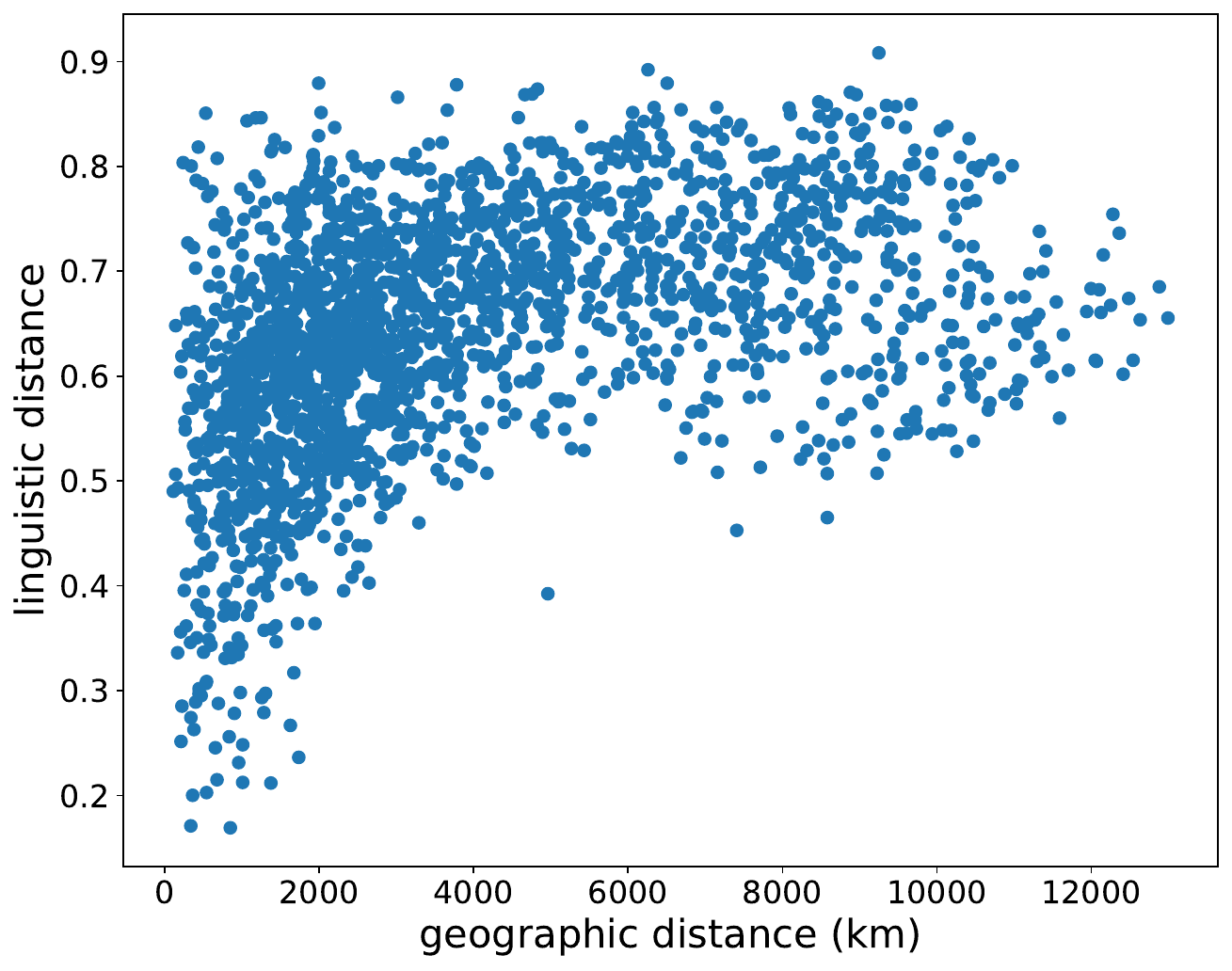}
 \caption{Linguistic and geographic distances between all language pairs considered. The distance correlation coefficient is $R_d=0.447$.}
 \label{fig:ling-geo_all}
\end{figure}

Finally, we compute the linguistic and geographic distances for all language pairs in order to globally explore the relation between these two variables. We show the resulting plot in Fig.~\ref{fig:ling-geo_all}. Quite generally, there is a positive correlation
between the syntactic and spatial variables. We can quantify this dependence with the the distance correlation coefficient $R_d$ \cite{10.1214/009053607000000505}. This coefficient is more general than Pearson's $R_p$ since it does not only measures linear dependence between variables. We find $R_d = 0.447$, with a p-value $<0.001$ calculated from a permutation test. Even though $R_d$ is not high due to the uncertainties associated to the noisy data and the geographical locations, the value is significantly greater than $0$ which indicates that geographical and linguistic distances are indeed correlated. 

\section{Conclusions}

In this work we have collected and analyzed parts of speech tagged sentences available in the Universal Dependencies library for $67$ contemporary languages located in a geographically contiguous region. Following an information-theoretic approach, we have provided evidence showing the effectiveness of utilizing the trigram probability distribution of parts-of-speech for characterizing the syntax statistics of languages.
Through this method, we have computed distances between languages by calculating statistical divergences between trigram distributions, revealing both well established language groupings and less familiar but already documented linguistic connections. 
Whereas most analyses of language families are conducted at the phonetic level, our syntactic approach, while yielding similar results, points to a robustness of linguistic family classifications across different linguistic levels. This opens up new avenues for linguistic research where syntactic data can complement phonetic data, providing a more comprehensive understanding of language evolution, classification and geographic relationships. We also stress that our approach is synchronic and thus complements the diachronic views more often encountered in historical linguistics.

A potential impact of our work is in the field of language documentation and revitalization. By quantifying syntactic similarities, our method can help identify which languages are most similar to endangered languages, therefore guiding efforts to develop educational materials and resources. For instance, if an endangered language has not been extensively documented, educational content from syntactically similar languages can be adapted more efficiently, preserving linguistic heritage more effectively.
Furthermore, our approach can be instrumental in computational linguistics, particularly in the development of multilingual natural language processing systems. Understanding syntactic similarities between languages can improve the performance of machine translation systems, especially for low-resource languages. Our results suggest that POS trigrams capture essential syntactic structures, which can be easily integrated into algorithms to enhance cross-linguistic transfer learning.
As another potential application, we mention that knowing the language distances across different levels can be useful in language teaching, as it may help focus on the aspects that differ most between the student's native language and the target language. 

Furthermore, our analysis has delved into the correlation between linguistic and geographic distances. We have found that spatially proximate languages tend to exhibit more similar morphosyntactic characteristics compared to those located farther apart. Our results suggest a logarithmic relation between these distances. This finding is in fact in good agreement with results reported in Refs.~\cite{nerbonne2010measuring,jager2018global}, where different measures of linguistic distances are defined. 
Due to the limitations of selecting a single point to represent the coordinates of an entire language, the results obtained can potentially be improved by considering more accurate methods to assign these locations, such as systematically selecting the regions where the languages are predominantly spoken, or by considering regional varieties that take into account the spatial variation of languages.

We emphasize that the methodology detailed in Sec.~\ref{sec:methods} for analyzing correlations within discrete sequences is versatile and applicable across numerous disciplines. These techniques hold relevance beyond the realm of linguistics and can be effectively employed in various fields, including but not limited to statistical physics, biology and data science, especially for systems that can be modeled as short-memory stochastic processes with few states.

A limitation of our research is that the UD dataset predominantly covers well documented languages. Therefore, less studied languages and dialects are underrepresented, which may limit the generalizability of our findings to the entire spectrum of global languages, since our results might not fully capture the syntactic diversity present in linguistic minorities. However, the UD library is frequently updated, 
regularly expanding existing corpora and incorporating new languages. It would be straightforward to extend our analysis to include more languages in future works.

Additionally, the use of POS tags, while effective for our analysis, might oversimplify the complexities of syntactic structures across different languages. POS tagging neglects many linguistic nuances and syntactic intricacies, which may lead to a loss of detailed information. This simplification can potentially affect the accuracy of our distance measurements, as some syntactic phenomena unique to specific languages may not be adequately represented. For example, languages with rich morphological systems or unique syntactic constructions may have their complexities reduced to basic categories that do not fully reflect their syntactic richness.
Future work should consider integrating more detailed morphosyntactic annotations, such as dependency and constituency parses, which are already included in the UD library, to capture a broader range of syntactic features.


\appendix

\renewcommand\thetable{\thesection.\arabic{table}} 
\setcounter{table}{0} 
\renewcommand\theHtable{Appendix.\thetable}

\section{List of languages}
\label{sec:appA}

We present in Table \ref{tab:languages} a list of the $67$ languages included in our study, categorized by language family, group, and morphological type. The latter is a simplified approximation that is included here for completeness. An exhaustive morphological analysis of the explored languages is beyond the scope of the present work.

\begin{longtable}{c|c|c|c}
  Family & Group & Language & Type \\
  \hline
   \multirow{3}{*}{Afro-Asiatic} & \multirow{3}{*}{Semitic} & {Arabic} &  {Isolating-Fusional} \\
                                 &                          & {Hebrew} & {Fusional} \\
                                 &                          & {Maltese} & {Fusional} \\
   
   \hline
   Austro-Asiatic & Viet-Muong & Vietnamese & Isolating \\
   \hline
   \multirow{2}{*}{Austronesian} & Javanese & Javanese & Agglutinative\\
   \cline{2-3} \cline{3-4}
   & Malayo-Sumbawan & Indonesian & Agglutinative\\
  \hline
  Basque & & Basque & Agglutinative\\
  \hline
  Dravidian & & Tamil & Agglutinative\\
  \hline
  \multirow{15}{*}{Indo-European} & Albanian & Gheg & Fusional\\
  \cline{2-3} \cline{3-4}
                & \multirow{2}{*}{Armenian} & Armenian & Agglutinative\\
              &             & Western Armenian & Agglutinative\\
 \cline{2-3} \cline{3-4}  
                & \multirow{2}{*}{Baltic}    & Latvian & Fusional\\
              &              & Lithuanian & Fusional\\
 \cline{2-3} \cline{3-4}
                & \multirow{5}{*}{Celtic}    & Breton    & Fusional\\
                &            & Irish     & Fusional\\
                &            & Manx      & Fusional\\
                &            & Scottish Gaelic & Fusional\\
                &            & Welsh            & Fusional\\
 \cline{2-3} \cline{3-4}                           
                & \multirow{9}{*}{Germanic}   & Afrikaans         & Isolating-Fusional\\
                &           & Danish            & Fusional\\
                &           & Dutch             & Fusional\\
                &           & English           & Isolating-Fusional\\
                &           & Faroese           & Fusional\\
                &           & German            & Fusional\\
                &           & Icelandic         & Fusional\\
                &           & Norwegian         & Fusional\\
                &           & Swedish           & Fusional\\
 \cline{2-3} \cline{3-4}
                & Greek     & Greek             & Fusional\\
 \cline{2-3} \cline{3-4}
                & \multirow{2}{*}{Indic}     & Hindi             & Fusional\\
                &           & Urdu              & Fusional\\
 \cline{2-3} \cline{3-4}
                & \multirow{2}{*}{Iranian}   & Kurmanji          & Fusional-Agglutinative\\
                &           & Persian           & Agglutinative\\
 \cline{2-3} \cline{3-4}
                & \multirow{7}{*}{Romance}   & Catalan           & Fusional\\
                &           & French            & Fusional\\
                &           & Galician          & Fusional\\
                &           & Italian           & Fusional\\
                &           & Portuguese        & Fusional\\
                &           & Romanian          & Fusional\\
                &           & Spanish           & Fusional\\
 \cline{2-3} \cline{3-4}
                &     & Belarusian        & Fusional\\
                &           & Bulgarian         & Isolating-Fusional\\
                &           & Croatian          & Fusional\\
                &           & Czech             & Fusional\\
                &           & Polish            & Fusional\\
                &           & Pomak             & Fusional\\
                & Slavic          & Russian           & Fusional\\
                &           & Serbian           & Fusional\\
                &           & Slovak            & Fusional\\
                &           & Slovenian         & Fusional\\
                &           & Ukrianian         & Fusional\\
                &           & Upper Sorbian     & Fusional\\
\hline
    Japanese      &           & Japanese            & Agglutinative\\
\hline
    Korean      &           & Korean            & Agglutinative\\
\hline
    Mongolic    &           & Buryat            & Agglutinative\\
\hline
    \multirow{2}{*}{Sino-Tibetan} &          & Cantonese         & Isolating\\
                &           & Chinese           & Isolating\\
\hline
    Tai-Kadai   &           & Thai              & Isolating\\
\hline
    Tungusic    &           & Xibe              & Agglutinative\\
\hline
    \multirow{3}{*}{Turkic}      & Northwestern & Kazakh         & Agglutinative\\
    \cline{2-3} \cline{3-4}
                & Southeastern  & Uyghur        & Agglutinative\\
    \cline{2-3} \cline{3-4}
                & Southwestern  & Turkish       & Agglutinative\\
\hline
    \multirow{3}{*}{Uralic}      & \multirow{2}{*}{Finnic}       & Estonian         & Agglutinative\\
                &               & Finnish          & Agglutinative\\
    \cline{2-3} \cline{3-4}
                & Mordvin       & Erzya             & Agglutinative\\
    \cline{2-3} \cline{3-4}
                & Permic        & Komi Zyrian       & Agglutinative\\
    \cline{2-3} \cline{3-4}
                & Sami          & North Sami        & Agglutinative\\
    \cline{2-3} \cline{3-4}
                & Ugric         & Hungarian         & Agglutinative\\

 \hline         
 \caption{List of languages included in this work, classified by language family, group and type.}
 \label{tab:languages}
\end{longtable}

\section{Predictability gain and block entropy}
\label{sec:appB}

The predictability gain of going from $u$th-order to $(u+1)$th-order transition probabilities, for $u\geq 1$, can be defined in terms of a conditional relative entropy \cite{cover} as
\begin{equation}
\mathcal{G}_u = \sum_{x_0,\ldots,x_{u+1}} p(x_0,x_1,\ldots,x_u,x_{u+1})\log\left(\dfrac{p(x_{u+1}|x_0,\ldots,x_u)}{p(x_{u+1}|x_1,\ldots,x_u)}  \right),
\label{eq:pg_2}
\end{equation}
where we adopt the notation that $p(x_0,\ldots,x_s) \equiv P(X_0=x_0,\ldots,X_s=x_s)$, with $x_i \in \lbrace z_j\rbrace_{j=0}^{L-1}$, and similarly for the transition probabilities. 

We will use the following general result for the difference between the block entropies $H_{r+1}-H_r$, demonstrated in the appendix of Ref.~\cite{juan}:
\begin{equation}
H_{r+1}-H_r = -\sum_{x_1,\ldots,x_{r+1}}p(x_1,\ldots,x_{r+1})\log(p(x_{r+1}|x_1,\ldots,x_r)).    
\label{eq:diffH}
\end{equation}

By the properties of the logarithm function, we can write Eq.~\eqref{eq:pg_2} as
\begin{equation}
\mathcal{G}_u = \sum_{x_0,\ldots,x_{u+1}} p(x_0,\ldots,x_{u+1})\log(p(x_{u+1}|x_0,\ldots,x_u)) - \sum_{x_0,\ldots,x_{u+1}} p(x_0,\ldots,x_{u+1})\log(p(x_{u+1}|x_1,\ldots,x_u)).    
\label{eq:pg_3}
\end{equation}
Adding $1$ to all indices in the first sum in Eq.~\eqref{eq:pg_3}, which is allowed given that we are considering homogeneous sequences, and using $\sum\limits_{x_0}p(x_0,x_1,\ldots,x_{u+1})=p(x_1,\ldots,x_{u+1})$, we write Eq.~\eqref{eq:pg_3} as
\begin{equation}
\mathcal{G}_u = \sum_{x_1,\ldots,x_{u+2}} p(x_1,\ldots,x_{u+2})\log(p(x_{u+2}|x_1,\ldots,x_{u+1})) - \sum_{x_1,\ldots,x_{u+1}} p(x_1,\ldots,x_{u+1})\log(p(x_{u+1}|x_1,\ldots,x_u)).  
\label{eq:pg_4}
\end{equation}
Using Eq.~\eqref{eq:diffH} we express Eq.~\eqref{eq:pg_4} as
\begin{align}
\begin{split}
\mathcal{G}_u = -(H_{u+2}-H_{u+1}) + (H_{u+1}-H_u) 
= -(H_{u+2}-2H_{u+1}+H_u),
\end{split}
\end{align}
which is exactly the result we wanted to prove. The case $u=0$ can be demonstrated similarly, considering that
\begin{equation}
\mathcal{G}_0 = \sum_{x_0,x_{1}} p(x_0,x_1)\log\left(\dfrac{p(x_{1}|x_0)}{p(x_{1})}  \right).
\label{eq:pg_0}
\end{equation}

\renewcommand\thefigure{\thesection.\arabic{figure}} 
\setcounter{figure}{0}  
\renewcommand\theHfigure{Appendix.\thefigure}

\section{Memory effects in POS sequences}
\label{sec:appC}

In Sec.~\ref{sec:motivation}, we have shown that POS sequences can be approximated as stochastic processes with memory 2 and that this approximation is good. However, this does not necessarily imply that the process has a memory $m$ exactly equal to 2. 
To see this, we compare the obtained value of $\mathcal{\hat{G}}_2$ with the ones calculated under the null hypothesis.
From the original dataset of language $\mathcal{L}$, composed of $R$ tagged sentences, we compute its second-order transition probabilities by means of Eq.~\eqref{eq:transprobs}.
We then generate $R$ sequences, each of equal length as the original sentences, from which we build the set $\lbrace \hat{n} \rbrace_1$.
We repeat this procedure $K$ times. 
By construction, the sets of observations $\lbrace \hat{n} \rbrace_1,\ldots,\lbrace \hat{n} \rbrace_K$, built from the generated groups of sequences, have the same size as the original set $\lbrace \hat{n} \rbrace$ of language $\mathcal{L}$ to ensure equal amount of data for comparison. Moreover, these numerical sequences have memory $m=2$ and consequently we expect the values $\mathcal{\hat{G}}_2[\lbrace \hat{n} \rbrace_k]$ to be close to $0$, for $k=1,\ldots,K$.
Therefore, we can estimate the p-value $p$ as
\begin{equation}
\hat{p} = \dfrac{\#\lbrace k:\mathcal{\hat{G}}_2[\lbrace \hat{n} \rbrace_k] \geq  \mathcal{\hat{G}}_2[\lbrace \hat{n} \rbrace] \rbrace}{K}.  
\end{equation}

\begin{figure}[t]
\sidesubfloat[]{\includegraphics[width=0.44\columnwidth]{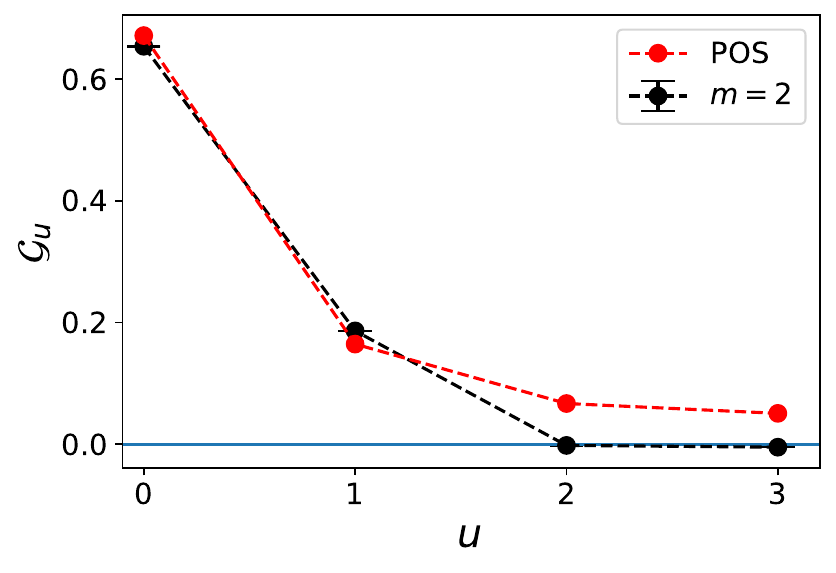}}
 \quad
\sidesubfloat[]{\includegraphics[width=0.44\columnwidth]{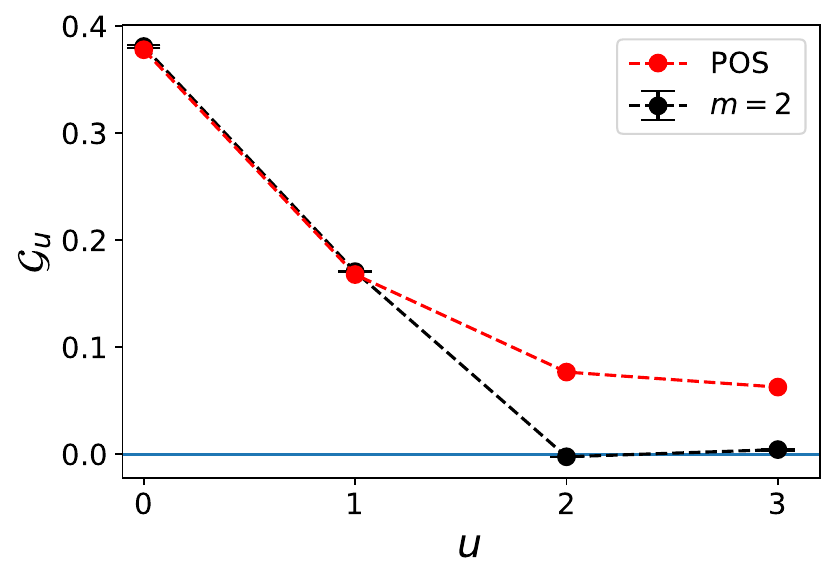}} \\
\sidesubfloat[]{\includegraphics[width=0.44\columnwidth]{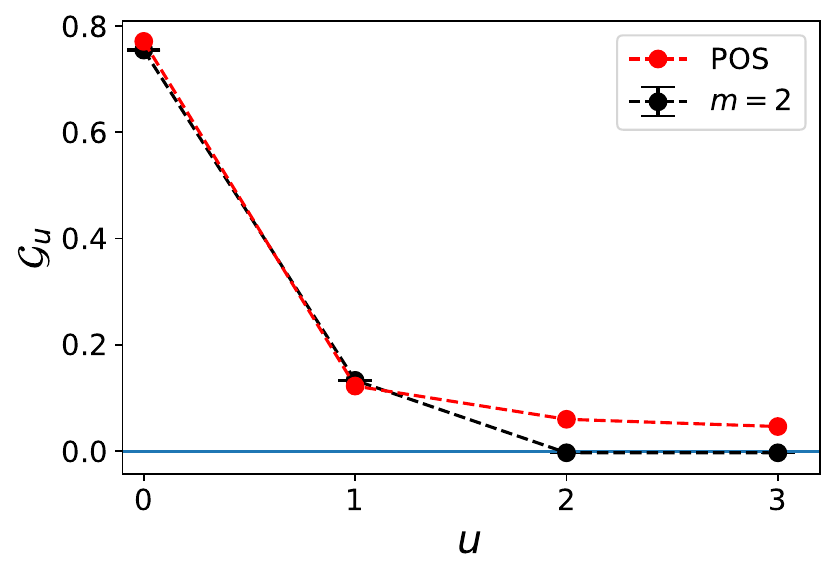}}
 \quad
\sidesubfloat[]{\includegraphics[width=0.44\columnwidth]{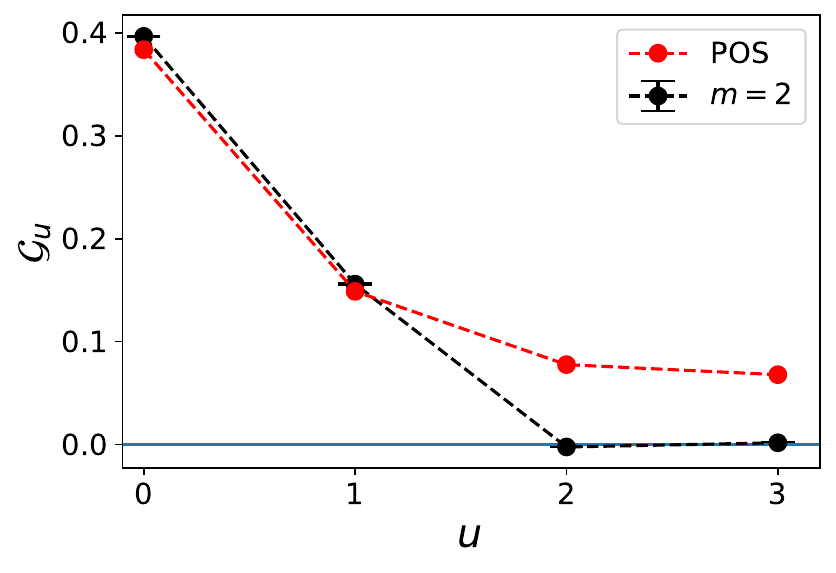}}
 \caption{Estimated predictability gain when considering $(u+1)$th-order instead of $u$th-order transition probabilities in sequences of POS for (a) German, (b) Icelandic, (c) Portuguese and (d) Czech, shown in red. The mean and standard deviation of the predictability gain estimated on numerically generated sequences of memory $2$, generated from the estimated transition probabilities of each language, are represented in black.}
 \label{fig:pg2}
\end{figure}

We apply this method to each of the four languages considered in Sec.~\ref{sec:motivation}, setting $K=1000$. We obtain $\hat{p} < 0.001$ for all cases, which leads us to reject the hypothesis that the POS sequences have memory exactly equal to $2$. For comparison, we plot in Fig.~\ref{fig:pg2} with black color the curves corresponding to the mean $\bar{\mathcal{G}}_u$ and standard deviation $s_u$ of the estimated predictability gain for the generated sequences of memory $2$, calculated as follows:
\begin{equation}
\bar{\mathcal{G}}_u = \dfrac{1}{K}\sum_{k=1}^{K} \hat{\mathcal{G}}_u[\lbrace \hat{n} \rbrace_k],    
\label{eq:mean_g}
\end{equation}
and 
\begin{equation}
s_u = \sqrt{\dfrac{1}{K-1}\sum_{k=1}^K \left(\hat{\mathcal{G}}_u[\lbrace \hat{n} \rbrace_k]-\bar{\mathcal{G}}_u\right)^2}.
\label{eq:std_g}
\end{equation}

We repeat the analysis with the hypothesis that $m=3$ and obtain similar results, indicating that the POS sequences for these languages possess a memory of at least $4$.
This is a possible hint of the observed long-range correlations in texts~\cite{altmann2012origin}. However, we stress that for the purposes of our work
the $m=2$ approximation works rather well.

\floatsetup[figure]{style=plain,subcapbesideposition=top}

\renewcommand\thefigure{\thesection.\arabic{figure}} 
\setcounter{figure}{0}  
\renewcommand\theHfigure{Appendix.\thefigure}

\section{Distance between texts belonging to a single language group}
\label{sec:appD}

\begin{figure}[t]
\sidesubfloat[]{\includegraphics[width=0.445\columnwidth]{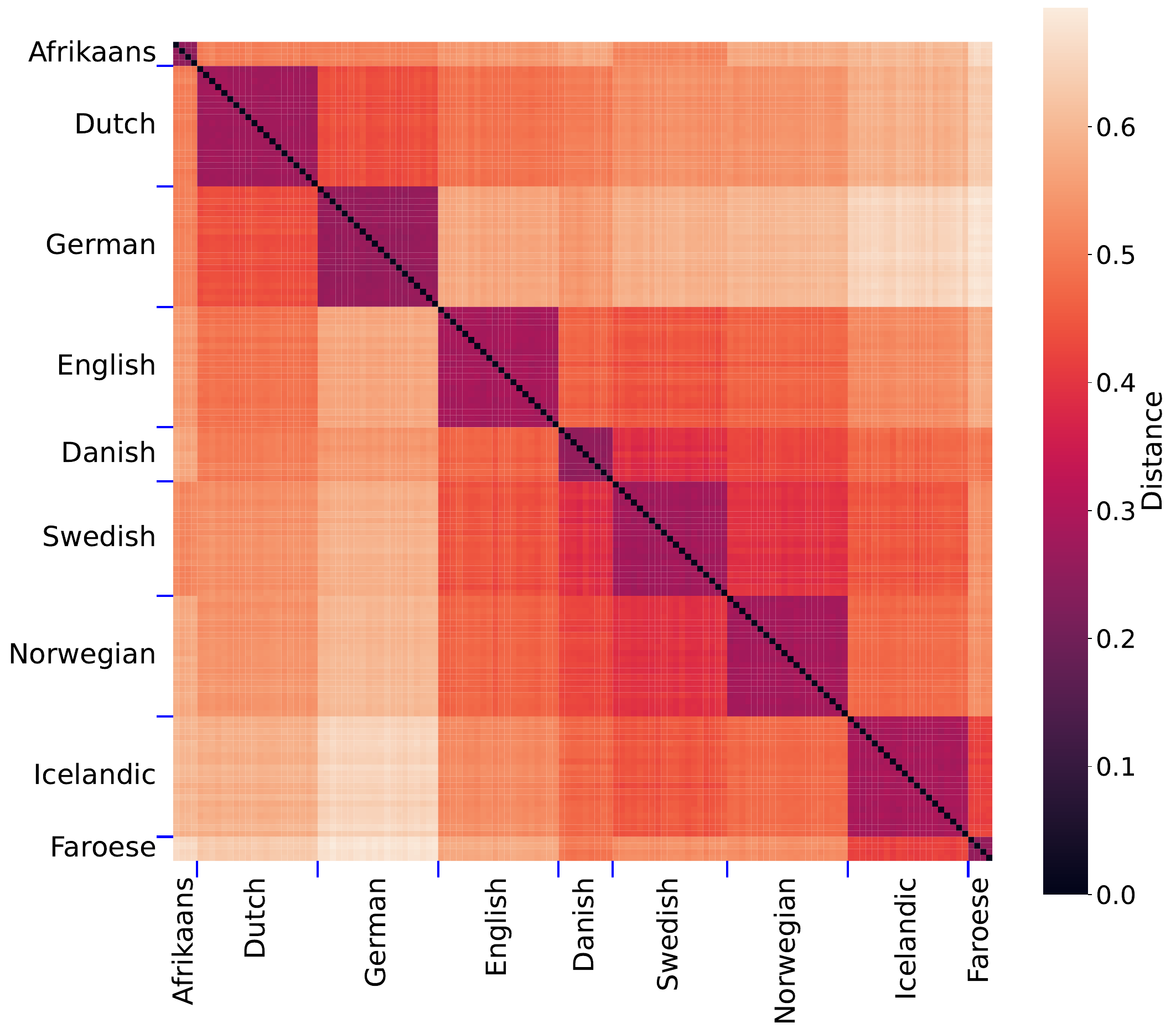}}
 \quad
\sidesubfloat[]{\includegraphics[width=0.445\columnwidth]{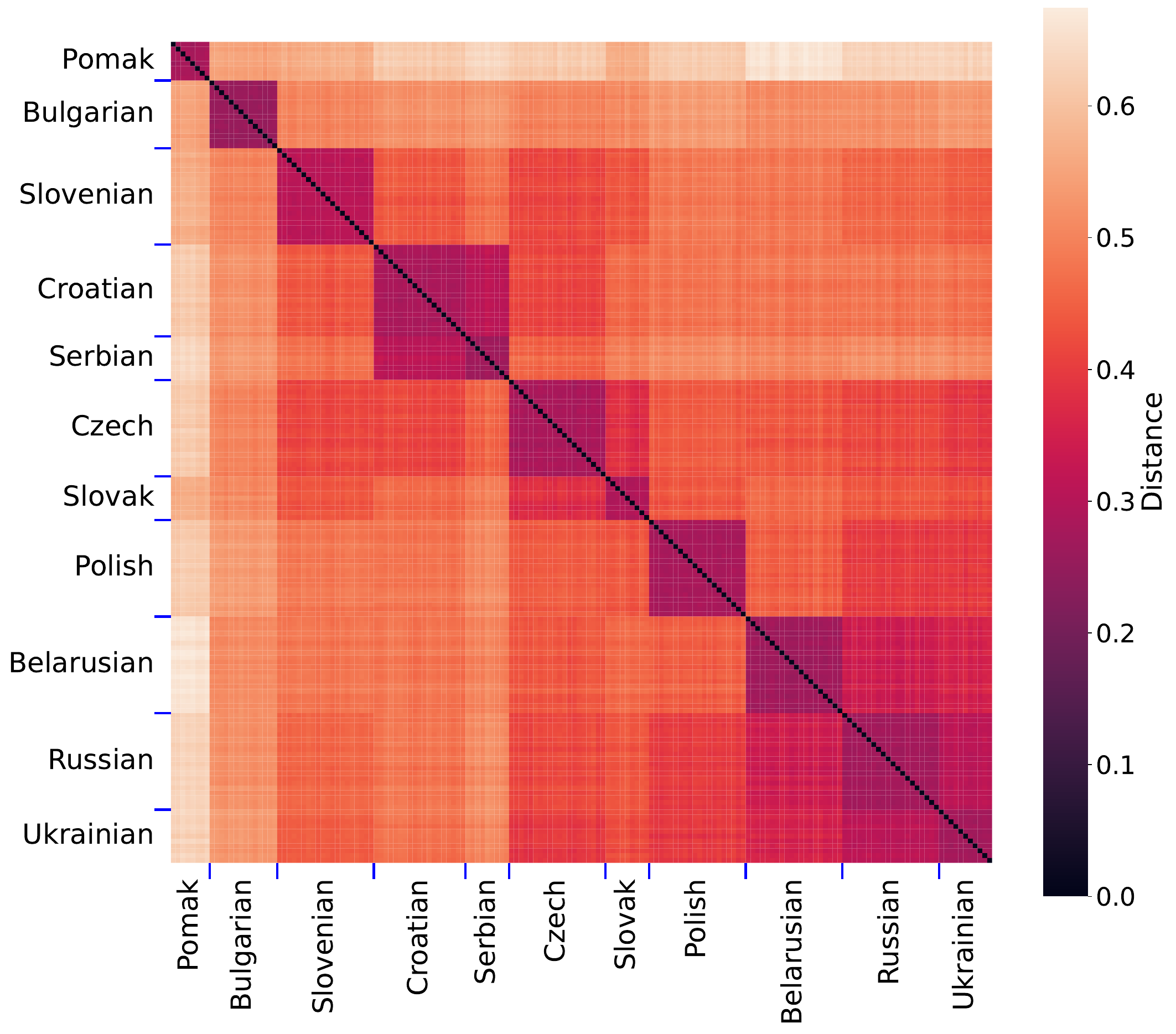}} 
 \caption{Heatmap representation of the pairwise Jensen-Shannon distances between portions of text of (a) Germanic and (b) Slavic languages.}
 \label{fig:families}
\end{figure}

We consider a single linguistic group. For each language in this group we randomly select tagged sentences from our database until we reach approximately $10^4$ tokens. We iterate this procedure a maximum of $20$ times, without replacement (for a few languages there is not enough data to perform this procedure $20$ times). 
Then, for each text portion of each language inside the given group we calculate its probability distribution of POS trigrams and compute the pairwise distances among texts with the JS metric. In Fig.~\ref{fig:families} we present heatmaps corresponding to the distance matrices obtained for Germanic (panel a) and Slavic (panel b) texts. We can observe that, in general, the distance between texts of the same language is smaller when compared to texts coming from distinct languages. This holds even for languages that are considered to be closely related, such as Croatian and Serbian, and Belarusian, Russian and Ukrainian.

\floatsetup[figure]{style=plain,subcapbesideposition=top}

This analysis demonstrates the reliability of our proposed metric to discern texts originated from the same language, which is generally a desirable result.

\renewcommand\thefigure{\thesection.\arabic{figure}} 
\setcounter{figure}{0}  
\renewcommand\theHfigure{Appendix.\thefigure}

\section{Clustermap obtained from the Hellinger distance matrix}
\label{sec:appHellinger}

Following the same procedure outlined in Sec.~\ref{sec:dist_matrix} of the main text, a clustermap was built from the Hellinger (H) distance matrix. The metric in this case is given by
\begin{equation}
    d_{_{H}}(\mathcal{L},\mathcal{L}') = \sqrt{\dfrac{1}{2} \sum_{j=0}^{L^{^{3}}-1} \left( \sqrt{p_{_\mathcal{L}}(b_j)} - \sqrt{p_{_{\mathcal{L}'}}(b_j)} \right)^2}.
    \label{eq:hell}
\end{equation}
Similarly to the JS divergence, $d_{_{H}}$ fulfills the requirements of a distance and lies in the $[ 0,1]$ interval.

\begin{figure}[t]
\includegraphics[width=1\textwidth]{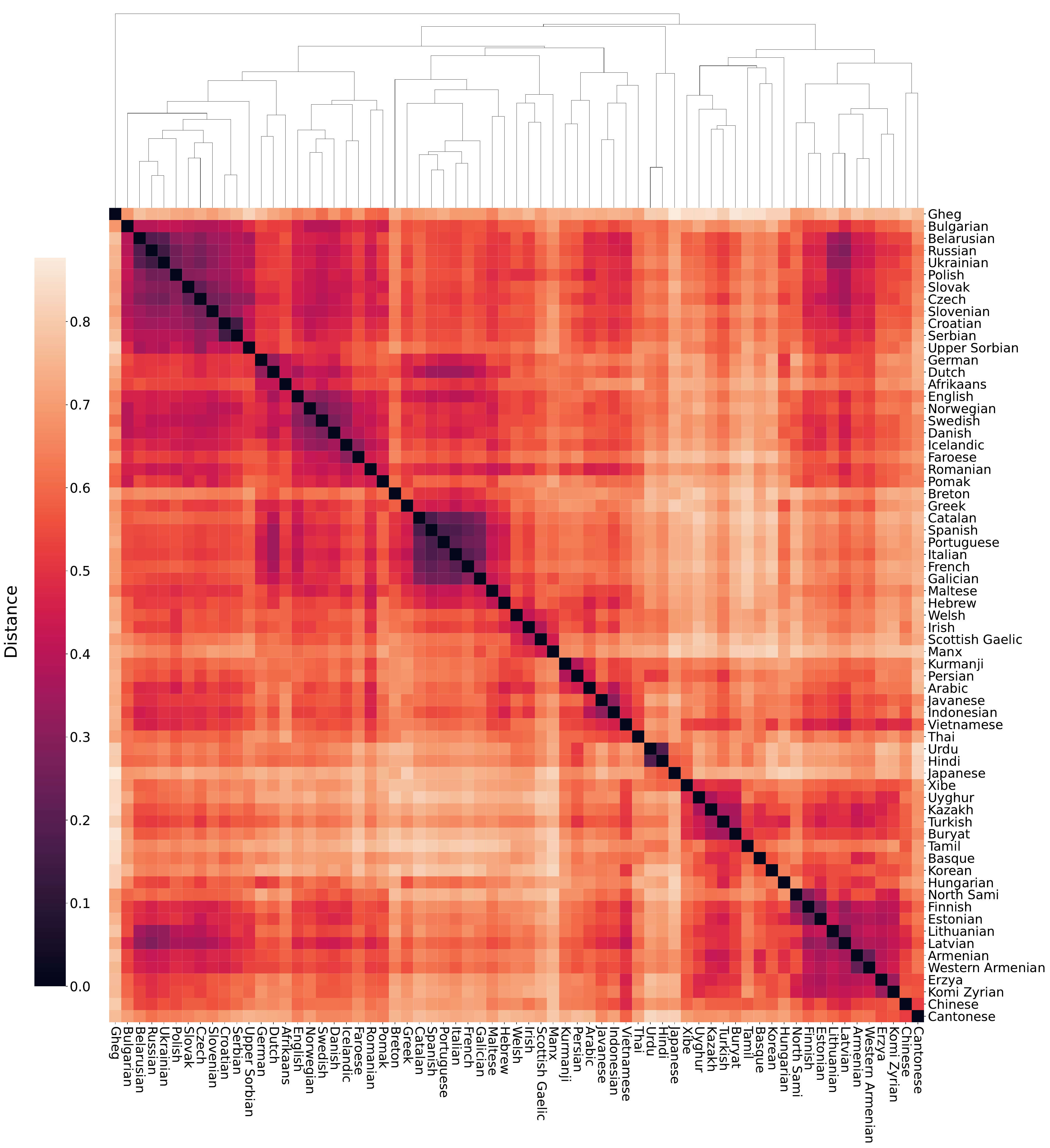}
 \caption{Clustermap constructed from the Hellinger distance matrix.}
 \label{fig:colormap_hellinger}
\end{figure}

Figure~\ref{fig:colormap_hellinger} shows that the constructed representation resembles the one obtained from the JS matrix. Not only the distributions of the colors match, but also the ordering of the rows and columns obtained with the hierarchichal clustering is similar. Only minor differences can be observed in the lower half of the colormaps but, nevertheless, one can identify the same cluster formations.

\renewcommand\thetable{\thesection.\arabic{table}} 
\setcounter{table}{0} 
\renewcommand\theHtable{Appendix.\thetable}

\clearpage

\section{Silhouette analysis and clusters obtained with $k$-medoids algorithm}
\label{sec:appF}

\begin{figure}[t]
\includegraphics[width=0.6\columnwidth]{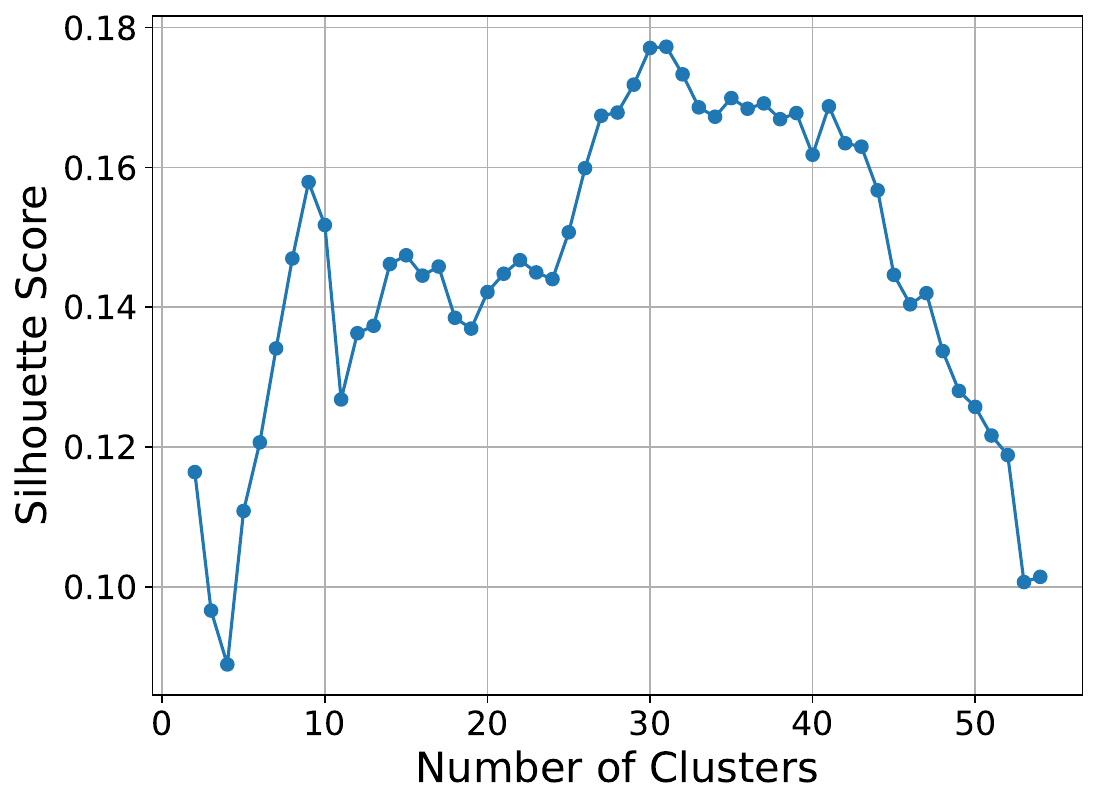}
 \caption{Silhouette score for various cluster numbers in a $k$-medoids clustering. The highest score is achieved with a cluster number of $31$.}
 \label{fig:silhoutte}
\end{figure}

In order to determine the optimal number of clusters for the $k$-medoids analysis from the JS distance matrix, we compute the silhouette score \cite{ROUSSEEUW198753} for various cluster numbers. The clusterization is then performed using the cluster number that yields the maximum silhouette score. We show the resulting plot in Fig.~\ref{fig:silhoutte}. As seen, the maximum occurs for $31$ clusters.

The clusters obtained using this method are presented in Table~\ref{tab:clusters}.

\begin{longtable}{c|c}
\fontsize{10}{11.4}\selectfont
Cluster & Languages \\
    \hline
    0 & Belarusian, Polish, Russia, Ukrainian, Latvian, Lithuanian \\
    1 & Catalan, French, Galician, Italian, Portuguese, Spanish \\
    2 & Buryat, Kazakh, Turkish, Uyghur  \\
    3 & Danish, English, Norwegian, Swedish \\
    4 & Hindi, Urdu \\
    5 & Estonian, Finnish, North Sami \\
    6 & Irish, Scottish Gaelic, Welsh \\
    7 & Arabic, Indonesian, Javanese \\
    8 & Bulgarian, Czech, Slovak, Slovenian, Upper Sorbian \\
    9 & Kurmanji, Persian \\
    10 & Armenian, Wester Armenian \\
    11 & Cantonese \\
    12 & Gheg \\
    13 & Japanese \\
    14 & Thai \\
    15 & Chinese \\
    16 & Tamil \\
    17 & Manx \\
    18 & Xibe \\
    19 & Korean \\
    20 & Afrikaans, Dutch, German \\
    21 & Faroese, Icelandic, Romanian \\
    22 & Erzya, Komi Zyrian \\
    23 & Hungarian \\
    24 & Basque \\
    25 & Hebrew, Maltese \\
    26 & Croatian, Serbian \\
    27 & Vietnamese \\
    28 & Breton \\
    29 & Pomak \\
    30 & Greek \\
    \hline
    \caption{Clusters formed with $k$-medoids algorithm from the Jensen-Shannon distance matrix.}
    \label{tab:clusters}
\end{longtable}

\renewcommand\thetable{\thesection.\arabic{table}} 
\setcounter{table}{0}
\setcounter{figure}{0}
\renewcommand\theHtable{Appendix.\thetable}

\section{Minimum spanning tree with tetragrams}
\label{sec:appG}

\begin{figure}[t]
\includegraphics[width=1\textwidth]{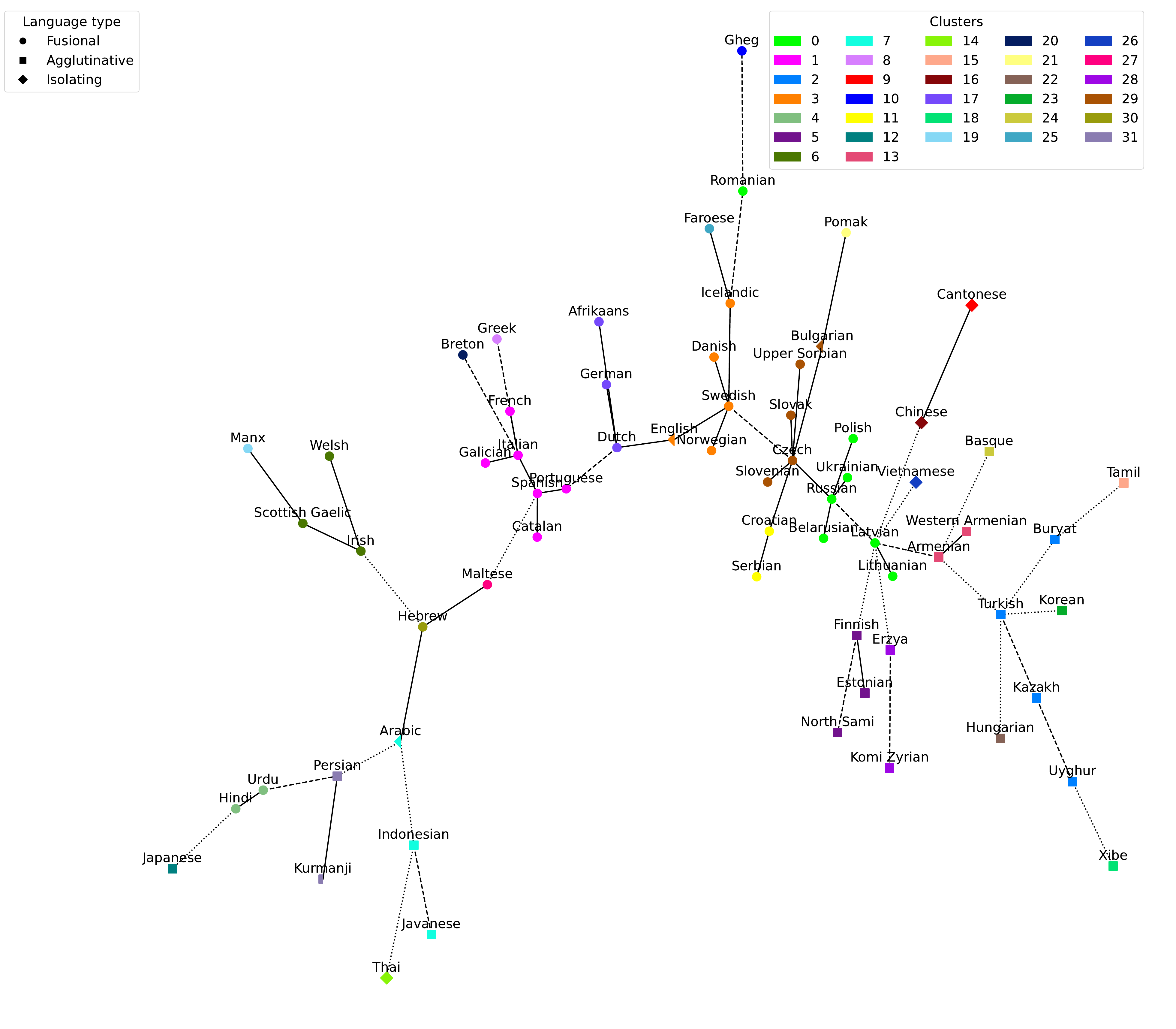}
 \caption{Minimum spanning tree generated from the Jensen-Shannon distance matrix, obtained from the probability distribution of POS tetragrams, with node colors representing clusters identified through $k$-medoids analysis. The shape of the nodes represents the language typology. Full lines are assigned to links between languages belonging to the same group; dashed lines for languages of the same family but different group; finally, dotted lines connect languages from distinct families.}
 \label{fig:mst_js_tetragram}
\end{figure}

In order to validate the results obtained in Sec.~\ref{sec:clusters} using trigrams, we perform a similar analysis using higher order $r$-grams
for the $67$ selected languages.

Specifically, in Fig.~\ref{fig:mst_js_tetragram} we depict the minimum spanning tree obtained using tetragrams, which can be compared with the tree shown in Fig.~\ref{fig:mst_js}.
For tetragrams, we find that the optimal number of clusters is $32$ (instead of $31$ for trigrams). We observe that the obtained clusters and connections in the tree are very similar in both cases.

One noticeable difference is that in Fig.~\ref{fig:mst_js_tetragram} there is a connection between Arabic and Hebrew that was missing in the tree of Fig.~\ref{fig:mst_js}. This is a nice finding since both languages are Semitic, although the $3$ languages of this family appear in $3$ different clusters.
This and other discrepancies can be attributed to small data samples, which can contain low values of tokens (of the order of ten thousand for certain languages), whereas the number of possible tetragrams is $15^4=50625$.

Choosing the size of the $r$-grams requires a compromise between the complexity of syntactic structures considered and the accuracy of the parameters estimation. In Sec.~\ref{sec:motivation} we have provided evidence that the choice of trigrams offers a good solution to this compromise and the analysis performed with tetragrams supports this claim, since there exist no significant differences with the results based on trigrams.


\section*{Availability of data and material}

The datasets of parts-of-speech for all languages analyzed during the current study are available in the Universal Dependencies repository at \url{http://hdl.handle.net/11234/1-5287}. The dataset of the geographic location of each language is available in the WALS repository at \url{https://wals.info/languoid}.

\section*{Acknowledgements}

We thank S.~J.~Greenhill for useful discussions.

\section*{Funding}

This work has been supported by the Agencia Estatal de Investigaci\'on (AEI, MICIU, Spain) MICIU/AEI/10.13039/501100011033 and Fondo Europeo de Desarrollo Regional (FEDER, UE) under Project APASOS (PID2021-122256NB-C21), the María de Maeztu Program for units of Excellence in R\&D, grant CEX2021-001164-M, and by the Government of the Balearic Islands CAIB fund ITS2017-006 under project CAFECONMIEL (PDR2020/51).

\bibliography{references}

\end{document}